%% file: ms.tex
\documentclass{article} % For LaTeX2e
\usepackage[accepted]{icml2024}

\usepackage{hyperref}
\usepackage{url}
\usepackage{graphicx}
\usepackage{algorithm}
\usepackage{algorithmic}
\usepackage{multirow}
\usepackage{subcaption}

\usepackage{amsfonts}       % blackboard math symbols
\usepackage{amsmath}
\usepackage{amssymb}
\usepackage{mathtools}
\usepackage{amsthm}
\usepackage{mathrsfs}

\usepackage{enumitem}
\usepackage{wrapfig}
\usepackage[export]{adjustbox}
\input{math_commands.tex}
\icmltitlerunning{Latent Noise Segmentation: the Emergence of Segmentation and Grouping with Latent Noise}

\begin{document}

\twocolumn[
\icmltitle{Latent Noise Segmentation: How Neural Noise Leads to the Emergence of Segmentation and Grouping}

%\maketitle
\icmlsetsymbol{equal}{*}

\begin{icmlauthorlist}
\icmlauthor{Ben Lonnqvist}{equal,yyy}
\icmlauthor{Zhengqing Wu}{equal,yyy}
\icmlauthor{Michael H. Herzog}{yyy}
\end{icmlauthorlist}

\icmlaffiliation{yyy}{École Polytechnique Fédérale de Lausanne (EPFL)}

\icmlcorrespondingauthor{Ben Lonnqvist}{ben.lonnqvist@epfl.ch}

\vskip 0.3in
]
\printAffiliationsAndNotice{\icmlEqualContribution}

\begin{abstract}
%suggested
Humans are able to segment images effortlessly without supervision using perceptual grouping.
Here, we propose a counter-intuitive computational approach to solving unsupervised perceptual grouping and segmentation: that they arise \textit{because} of neural noise, rather than in spite of it. 
We (1) mathematically demonstrate that under realistic assumptions, neural noise can be used to separate objects from each other; (2) that adding noise in a DNN enables the network to segment images even though it was never trained on any segmentation labels; and (3) that segmenting objects using noise results in segmentation performance that aligns with the perceptual grouping phenomena observed in humans, and is sample-efficient. 
We introduce the Good Gestalt (GG) datasets --- six datasets designed to specifically test perceptual grouping, and show that our DNN models reproduce many important phenomena in human perception, such as illusory contours, closure, continuity, proximity, and occlusion. 
Finally, we (4) show that our model improves performance on our GG datasets compared to other tested unsupervised models by $24.9\%$.
Together, our results suggest a novel unsupervised segmentation method requiring few assumptions, a new explanation for the formation of perceptual grouping, and a novel potential benefit of neural noise.

\end{abstract}

\section{Introduction and Related Work}

Humans perceptually group object parts in scenes \citep{wagemans2012a, wagemans2012b, jakel2016, roelfsema2006}, which allows them to interpret and segment elements in context, even when they deviate from stored templates \citep{geirhos2019, herzog2015, holcombe2001}. 
For example, the black and white stripes of a zebra are perceptually grouped together despite their vastly different colors.
Historically, perceptual grouping in humans has been studied by the Gestaltists, who view the Gestalt principles of grouping (such as grouping by proximity and continuity) as crucial computational principles. 
This human capability to perform perceptual grouping is in stark contrast to object recognition in Deep Neural Networks (DNNs), which have been shown to struggle with both robustness \citep{geirhos2018, madry2018, moosavi-dezfooli2016} and grouping \citep{bowers2022, linsley2018, doerig2020, biscione2023}. 
Understanding the underlying mechanisms of perceptual grouping is critical for better modeling of human vision, and a more fundamental understanding of segmentation in general.

\begin{figure*}[h]
%\framebox[4.0in]{$\;$}
\begin{center}
\includegraphics[scale=0.32]{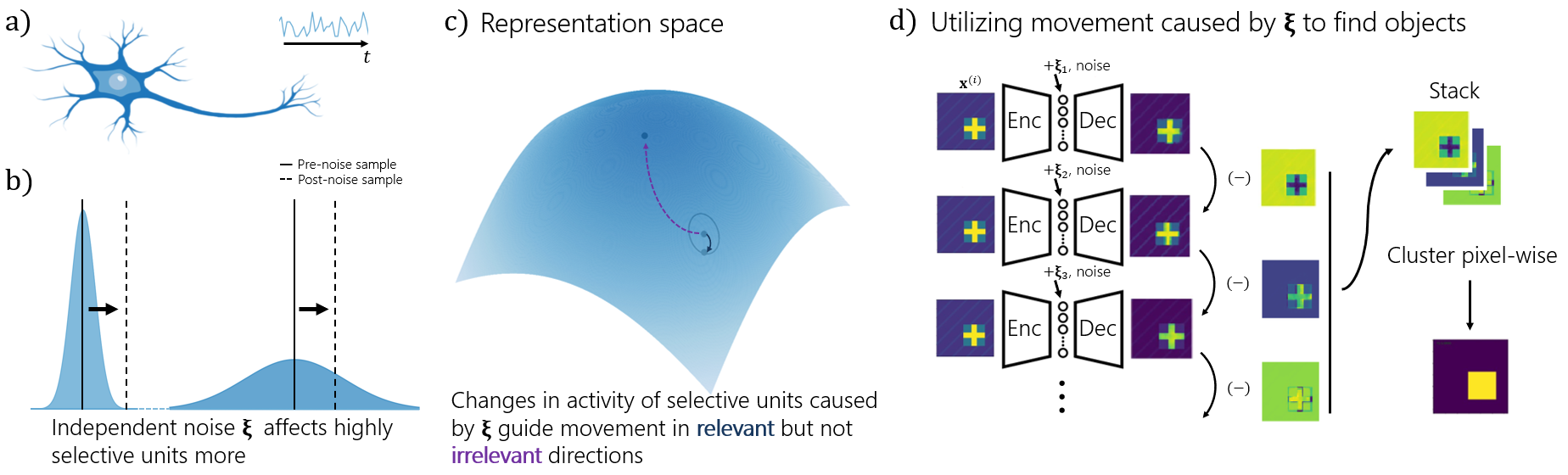}
\end{center}
\caption{\textbf{Latent Noise Segmentation Schematic Illustration.} \textbf{(a)} Biological neurons are highly noisy. For example, thermal noise and ion channel shot noise \citep{manwani1998} contribute to independent noise in neurons. \textbf{(b)} Independent noise affects the output of neurons that are highly selective to a stimulus feature (\textit{left}) than neurons that are less selective (\textit{right}). Solid lines indicate the mean of the noise-free activity distribution, and dashed lines indicate the actual sample after independent noise is added. \textit{The x-axis in the illustration is not meaningful.} \textbf{(c)} In the system’s representational space (as indicated by the surface where input images are mapped to points on that surface), the changes caused by independent noise cause meaningful changes to the model’s representation in relevant directions, e.g., local Principal Component (PC) directions, but not irrelevant directions (that would substantially change the model's representation of the input). \textbf{(d)} This yields information about the objects in the input image, and can be used to segment the input images. An input image $\mathbf{x}^{(i)}$ is fed to an autoencoder network, and noisy samples are drawn and consecutively subtracted from each other. These outputs contain information about the changes induced by noise in latent space, cast into image space. The outputs are stacked and clustered pixel-wise to generate a segmentation mask.}
\label{fig:illustration}
\end{figure*}

While incorporating explicit or implicit segmentation and grouping mechanisms in DNNs has seen substantial rise in interest in recent years \citep{greff2020, hinton2023, linsley2018, sabour2017}, mainstream instance segmentation approaches often either train models directly to segment using labeled samples \citep{ronneberger2015, long2015, kirillov2023}, or inject a substantial number of biases into the model architecture \citep{hamilton2022, engelcke2022, engelcke2020genesis, wang2022freesolo}.
The current prevailing approach is the implementation of object slots in the model architecture \citep{locatello2020}, which maintains information about object instances in a separately maintained vector.
Slot-based approaches are conceptually limited by the fact that the number of slots is fixed in the model architecture and cannot be adapted by learning, nor on an image-by-image basis.
%Furthermore, due to the added complexity in model architecture, such models are slow to train \citep{lowe2022}.
An approach that attempts to remedy the problems above was proposed by \citet{lowe2022}, who suggest a neuroscience-inspired solution to object learning: the object identities are stored in phase values of complex variables, which mimics the temporal synchrony hypothesis of brain function \citep{milner1974}.
While the approach can handle a variable number of objects, it is limited in the maximum number of objects it can represent (but see \citet{stanic2023}).
To date, there is no unified and simple principle that allows DNNs to perform segmentation and grouping in a number of different, seemingly unrelated contexts with few \textit{in-principle} limitations. 
In this work, we seek to remedy this gap.

Specifically, we develop a direct test benchmark of a network’s segmentation and grouping capability by requiring it to segment images in a way that necessitates perceptual grouping –-- furthermore, the network must do this without ever being trained on the stimuli in question, and without ever being trained on the task of segmentation, similarly to how the human visual system is not.
To simultaneously accomplish all these goals, we posit a seemingly counterintuitive idea that we call Latent Noise Segmentation (LNS):

\quad \textbf{\textit{Neural noise enables deep neural networks to perceptually group and segment.}}

See Figure \ref{fig:illustration} for a schematic overview of the approach. In brief, we show that by injecting independent noise to a hidden layer of a DNN, we can turn DNNs trained on generic image reconstruction tasks to systems capable of perceptual grouping and segmentation.
%We show that by injecting activity-independent noise to a hidden layer of a DNN, we can turn DNNs trained on generic image reconstruction tasks to systems capable of perceptual grouping and segmentation. 
%We apply our method to six novel synthetic datasets inspired by a century of Gestalt psychology and the study of grouping in humans \citep{wertheimer1923, koffka1922}, and show that without any further assumptions, simply adding noise to the latent layer of an autoencoder is sufficient to reproduce the main results of the hypothesized principles of perceptual grouping \citep{wagemans2012a, wagemans2012b, todorovic2008}.

In the following sections, we describe the general setting of LNS and its specific implementation in an autoencoder DNN (Section \ref{section:model_and_methods}).
We mathematically demonstrate that under certain (relatively loose) assumptions from basic optics, the presence of latent noise allows visual entities, which do not perfectly co-vary with each other, to be separated into individual objects (Section \ref{section:latent_noise_segmentation} and Appendix \ref{appendix:proof}). 
We develop a comprehensive test benchmark of segmentation tasks inspired by a century of Gestalt psychology and the study of grouping in humans \citep{wertheimer1923, koffka1922} that we call the Good Gestalt (GG) datasets.
We empirically show that LNS directly reproduces the result of the elusive Gestalt principles of grouping in segmentation masks (Section \ref{section:GG}). 
Finally, we study how practically feasible LNS is (that is, how in-principle usable the method is by any system constrained by limitations like compute time or noise magnitude, such as the primate visual system) by evaluating how segmentation performance varies with different model learning rules, noise levels, and the number of time steps the model takes to segment (Section \ref{section:results}). 
Our results suggest that a practically feasible number of time steps (as few as a handful) are sufficient to reliably segment and that while encouraging a prior distribution in a model does not improve its segmentation performance, it stabilizes the optimal amount of noise needed for segmentation across all datasets and correctly identifies the appropriate number of objects.

\section{Model and Methods}
\label{section:model_and_methods}
%Here, we describe the general setting of our methodology, how we obtain segmentation masks with LNS, the design of our datasets, and model evaluation metrics.

\textbf{Basic setup.} Our model architecture consists of two primary components. 
The backbone of the model is a pre-trained (Variational) Autoencoder (VAE; \citet{kramer1991, kingma2013, higgins2017}), where an encoder $\operatorname{Enc}(\mathbf{x}) = q_\phi (\mathbf{z} | \mathbf{x})$ learns a compressed latent distribution, and a decoder $\operatorname{Dec}(\mathbf{z}) = p_\theta (\mathbf{x} | \mathbf{z})$ models the data that generated the latent representation $\mathbf{z}$.
In practice, the model is optimized using the Evidence Lower Bound (ELBO):
\begin{equation}
    ELBO=\mathbb{E}_{q_\phi (\mathbf{z}|\mathbf{x})}[\log p_\theta(\mathbf{x}|\mathbf{z})]- \beta D_{KL}(q_\phi (\mathbf{z}|\mathbf{x})||p(\mathbf{z})),
\end{equation}
where $D_{KL}$ is the Kullback-Leibler divergence, and $p(\mathbf{z})$ is the prior set to $\mathcal{N}(0, \mathbf{I})$. 
$\beta$ is a configurable coefficient facilitating reconstruction quality, which we set automatically through the GECO loss \citep{rezende2018}. 
For a discussion about alternate architectures, see Appendix \ref{app:proof_discussion}.

%In our work, we use GECO to slowly increase $\beta$ from a small value to $>1$ \textcolor{red}{(is this important)} throughout training, facilitating good reconstruction quality and encouraging disentanglement.
%On top of this trained model, we build a noisy segmentation process which enables the extraction and combination of information from the model’s latent space to generate segmentation masks.
%Full model details can be found in Appendix \ref{appendix:implementation}.

\begin{algorithm}
\caption{Latent Noise Segmentation.}\label{alg:noise}
\begin{algorithmic}[1]
\REQUIRE image $\mathbf{x}^i$, time steps $N$, small noise variance $\sigma_{small}^2$
\REQUIRE trained $(\beta)$-VAE with $\operatorname{Enc}_{\phi}, \operatorname{Dec}_{\theta}$
\STATE \textbf{Procedure} \textsc{Segment}($\mathbf{x}^i, {N}$)
\STATE $\boldsymbol{\mu}(\mathbf{x}^i) \gets \operatorname{Enc}(\mathbf{x}^i)$ %\COMMENT{Get latent unit means}
\FOR{$n=1, \dots, N$}
    \STATE $\boldsymbol{\xi}_{n} \gets \mathcal{N}(0, \sigma_{small}^2)$ %\COMMENT{i.i.d. Gaussian noise}
    \STATE $\Tilde{\mathbf{x}}_n^i \gets  \operatorname{Dec}\left(\boldsymbol{\mu}\left(\mathbf{x}^{i}\right)+\boldsymbol{\xi}_{n}\right)$%\COMMENT{Save perturbed decoder output}
    \IF{$n\geq 2$}
        \STATE $\Delta \Tilde{\mathbf{x}}_{n-1}^i \gets \Tilde{\mathbf{x}}_n^i - \Tilde{\mathbf{x}}_{n-1}^i$ %\COMMENT{Subtract outputs}
    \ENDIF
\ENDFOR
\STATE Let $\Delta \Tilde{\mathbf{X}}^i \equiv [\Delta\Tilde{\mathbf{x}}_{1}^i, \Delta\Tilde{\mathbf{x}}_{2}^i, \ldots, \Delta\Tilde{\mathbf{x}}_{N-1}^i]$
\COMMENT{$\Delta\Tilde{\mathbf{X}}^i$ has dimension $img_x \times img_y \times c \times (N-1)$}
\STATE Cluster $\Delta \Tilde{\mathbf{X}}^i$ %\COMMENT{Separate pixel identities} 
\STATE Assign segmentation mask values to pixels according to clustering
\STATE \textbf{End Procedure}
\end{algorithmic}
\end{algorithm}

\subsection{Latent Noise Segmentation}
\label{section:latent_noise_segmentation}
%Our method of segmentation is simple both conceptually and in practice. 
After training the model backbone, we build a noisy segmentation process which enables the extraction and combination of information from the model’s latent space to generate segmentation masks.
To do this, we add i.i.d. noise $\boldsymbol{\xi}_n$ to the latent variables $\mathbf{z}$.
The process is repeated $N$ times, allowing us to obtain information about which objects exist in the model outputs.
For every time step, two consecutive model outputs are subtracted from each other to find what in the output images $\Tilde{\mathbf{x}}$ \textit{changed with respect to changes in latent space} $\mathbf{z}$.
At the end of the process, the stack of subtracted outputs is clustered using Agglomerative Clustering \citep{scikit-learn}, resulting in a segmentation mask.
Full details of the algorithm are shown in Algorithm \ref{alg:noise}, a schematic is shown in Figure \ref{fig:illustration}, and additional details can be found in Appendix \ref{appendix:implementation}.

%The goal of our segmentation process is to extract and combine object-related information in latent space and to map it to image space by adding activity-independent noise to the samples of the latent layer's activations.
\subsection{The Role of Independent Noise}

Latent Noise Segmentation works due to several key reasons.
Independent noise $\boldsymbol{\xi}$ induces the locally monotone mapping $f: \Delta \mathbf{z} \rightarrow \Delta \Tilde{\mathbf{x}}$ from the latent variables to the output learned by the model.
A potential problem is that this mapping happens for all units --- even those that are not meaningful for the task of segmentation --- and as such, they might adversely affect segmentation performance.
To solve this, one could identify the meaningful units by perturbing them individually (like in latent traversal \citep{kingma2013, higgins2017}, but not only is this computationally expensive and impractical, it is also biologically implausible. 

The solution to the problem comes from the intuition that units that meaningfully code for relevant stimulus features have a substantially higher derivative $\Delta \Tilde{\mathbf{x}}=\Delta \mathbf{z} * \frac{\delta \Tilde{\mathbf{x}}}{\delta \mathbf{z}}$ in the neighborhood of the stimulus $\mathbf{x}$.
This is because for the decoder to achieve its goal of low reconstruction loss, it should be sensitive to small changes in the activity of units that are coding for relevant features in $\mathbf{x}$, but not for irrelevant features.
This means that we can perturb all units simultaneously with independent noise: training on a generic reconstruction loss encourages the network to code in a manner that allows independent noise to reveal the relevant derivative direction, making the representation clusterable for segmentation (Figure \ref{fig:illustration}b).
Indeed, the goal of independent noise is not to perform segmentation itself, but to reveal the local neighborhood of the input stimulus and thus enable segmentation, \textit{from the perspective of the decoder}.
\begin{figure*}
\begin{center}
%\framebox[4.0in]{$\;$}
\includegraphics[scale=0.40]{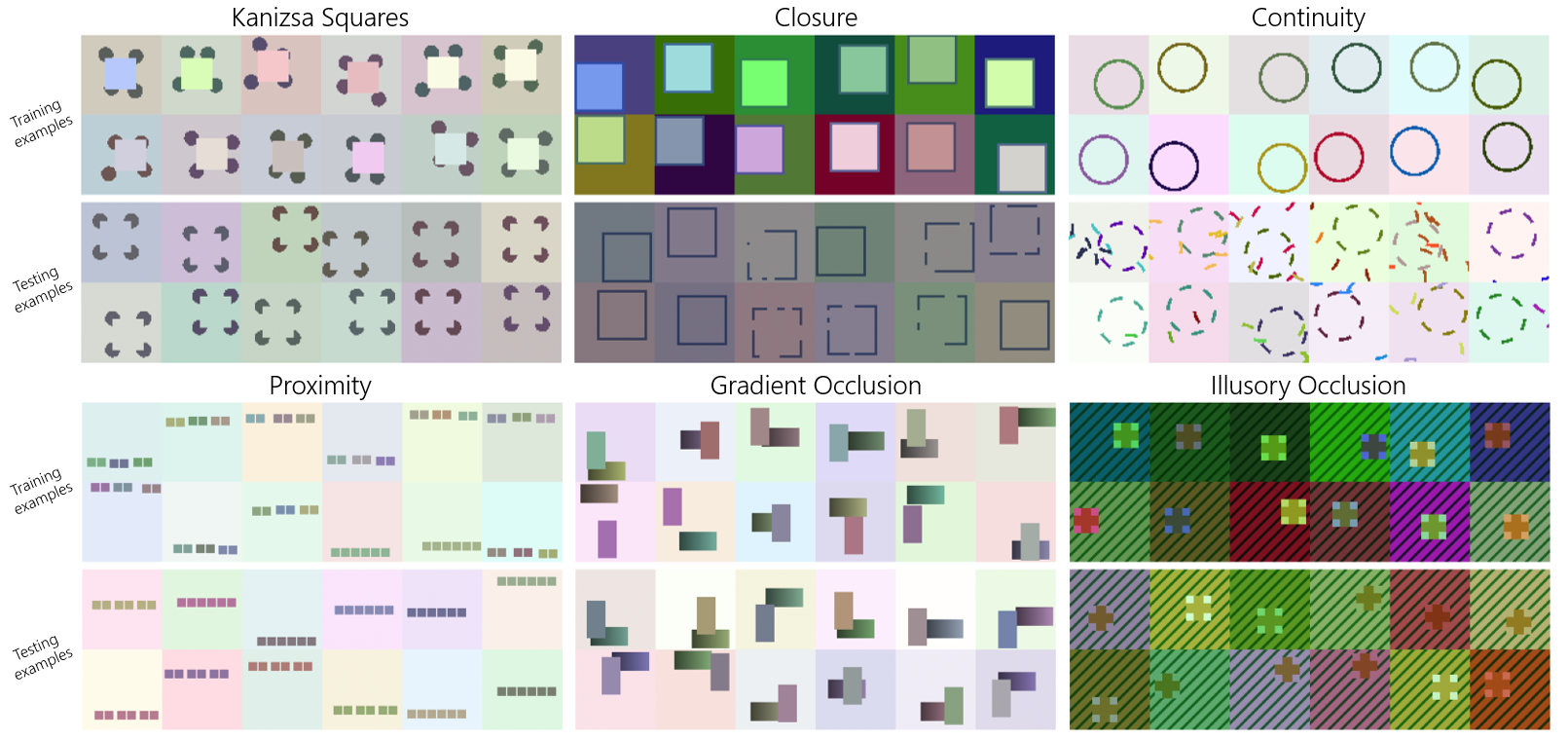}
\end{center}
\caption{\textbf{The Good Gestalt (GG) Datasets.} Zoom in for an ideal viewing experience. 
The first two rows of images show training image examples, while the second two rows show images of testing examples.}
%\textbf{Kanizsa Squares}: The training set consists of four circles occluded by a foreground square. In the testing set, the circles are aligned and their corners are cut, so that the appearance of an occluding square is formed. \textbf{Closure}: The training set contains squares with borders. The testing set contains borders that either completely or partially enclose a square-shaped region. \textbf{Continuity}: The training set consists of circles. In the testing set, the circles are cut into segments, of approximately $50\%$ density, and additional cluttering line segments are added. \textbf{Proximity}: In the training set either a set of six squares is present, or three sets of two squares of different colors and positions. In the testing set, the three sets of two squares are placed on the same horizontal plane and have the same color, leaving small offsets as the only cue to separate them. \textbf{Gradient Occlusion}: In both the training and testing sets, a background element made up of a gradient of colors is partially occluded by a foreground object. \textbf{Illusory Occlusion}: In the training set, a foreground object consisting of four squares and a cross occludes a background consisting of a striped pattern. In the testing set, either the squares or the cross of the foreground object is set to the color value of either the static background or the stripes.}
\label{fig:examples}
\end{figure*}
We provide a more complete mathematical intuition of exactly why noise perturbation on latent units causes a meaningful object segmentation in the case of the VAE in Appendix \ref{appendix:proof}.
In summary, we observe that noise induces the output pixels belonging to the same part of the image to co-vary with each other, while pixels belonging to different parts do not.
Following a principal component argument \citep{demystifying_inductive_biases, VAE_pursues_pca}, we show that the derivatives of the decoder's output with respect to the latent units are the local principal component directions of the training data.
Together, these phenomena enable meaningful object segmentation.

\subsection{The Good Gestalt (GG) Datasets}
\label{section:GG}

To evaluate our models, we developed the six datasets that we together call the Good Gestalt (GG) datasets.
The notion of Gestalt comes from the study of perception \citep{wertheimer1923, jakel2016, wagemans2012a, wagemans2012b}, and is perhaps best summarized as `the whole is greater than the sum of its parts`.
The GG datasets aim to address a large variety of different rules of perceptual organization, otherwise known as Gestalt principles of grouping like continuity, closure and proximity --- here, we outline the reasons for studying machine perception using such datasets, and then describe the specifics of our datasets.
Example stimuli of the training and testing samples are shown in Figure \ref{fig:examples}.

\textbf{Reasons for studying Gestalt.}
In almost all the Gestalt stimuli we study in this work, we assume that in training the luminances (pixel values) of different objects in a visual scene are independent, while the luminances (pixel values) of the same object co-vary.
This assumption follows a simplified view on optics (for details, see Appendix \ref{app: gg dataset details}), and is important as some principles of perceptual grouping have been shown to be consistent with the statistical structure of the natural environment \citep{elder2002, geisler2001, sigman2001}.
In testing, the model should generalize this information in ambiguous contexts, where the pixel values of the images are not fully informative of the object.
For example, in the Kanizsa Squares dataset (Figure \ref{fig:examples}, \textit{Kanizsa Squares, Testing examples}), the pixel value of the background is the same as the pixel value of the illusory square in the middle of the cornering circles.
%This puts a crucial tension on models: the more veridically the model perceives the stimulus, the less it can rely on the actual contents of the input image to make a correct judgment about the objects in the image.
%This tension is critical, as humans are able to perceive, for example, such illusory squares perfectly well, even though the square does not veridically exist.

%The tension between the actual stimulus contents and the desired model output is what allows us to keep the stimuli simple while keeping the task difficult.
%Because of that, we have a substantial amount of control over where the ability to segment such inputs comes from: past experience.

%In other words, we can precisely control our stimulus setting such that an appropriate model of Gestalt can generalize to perceiving difficult test cases only by assuming that it has perceived some ecologically plausible basic shapes, like squares, before.

\begin{figure*}
\begin{center}
%\framebox[4.0in]{$\;$}
\includegraphics[scale=0.35]{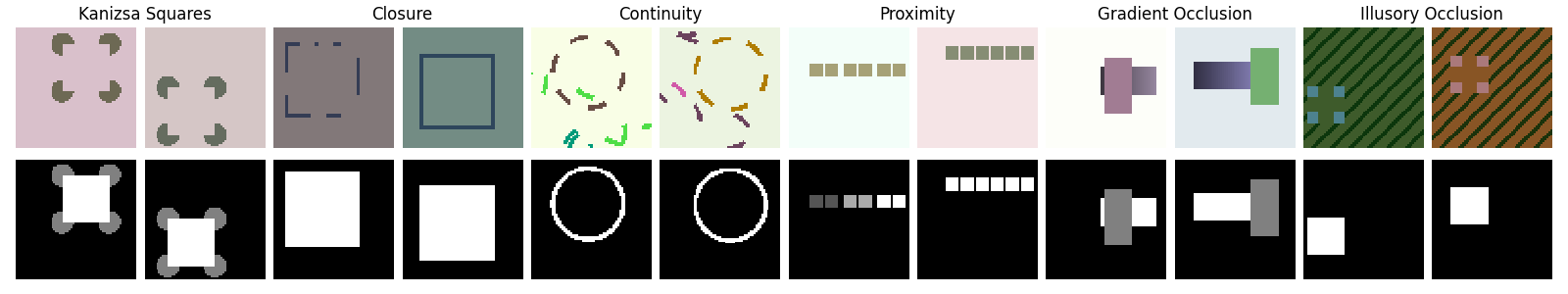}
\end{center}
\caption{\textbf{Target segmentation mask examples of the GG datasets.} The first row of images shows training image examples, while the second row shows images of testing examples. Different colors indicate different objects. \textbf{Kanizsa Squares}: The model should segment a square in the center, and four background circles separately from the background. \textbf{Closure}: The model should segment a square. \textbf{Continuity}: The model should segment a circle traced by the relevant line segments to ``complete the circle". \textbf{Proximity}: The model should segment a set of six squares together, or three sets of two squares when the proximity cue is given. \textbf{Gradient Occlusion}: The model should segment the two rectangles separately. \textbf{Illusory Occlusion}: The model should segment the static background and stripes together, and the foreground object parts together.}
\label{fig:segmasks}
\end{figure*}

We show target segmentation masks for the GG datasets in Figure \ref{fig:segmasks}.
To determine what these segmentation masks should be, our focus here is not about \textit{what} specifically humans perceive in a specific psychophysical experiment, but rather about what humans \textit{can} perceive given the right instruction.
%Perceiving Gestalt laws in such simple stimuli is very obvious --- we encourage the reader to zoom in on Figures \ref{fig:examples} and \ref{fig:segmasks} and to examine the examples. 
%Indeed, the question here is not about \textit{what} specifically humans perceive in a specific psychophysical experiment, but rather about what humans \textit{can} perceive given the right instruction.
This is part of what makes studying Gestalt so difficult: simple phrasing about what a participant should segment in such an image can potentially crucially affect how they segment the image.
This is why in our GG datasets, we did not collect human data, but instead focus on the generic question of ``which elements in the image belong together?", and defined the target segmentation masks for the test set evaluation on the basis of this question, focusing on evidence from human research in general from the past century \citep{wertheimer1923, kimchi1992}.

In total, the GG datasets are made up of six individual datasets, for which we explain the main premise individually below.
More information on the datasets can be found in Appendix \ref{app: gg dataset details}.

\begin{description}[itemsep=0pt]
  \item[Kanizsa Squares]to study the perception of illusory figures. Humans are able to perceive illusory contours or illusory objects when cued by, for example, Pacman-like circles \citep{wang2012, lesher1995}.
  \item[Closure]to study whether a model can combine individual image elements to form an object. The human ability to group objects together when they form a closed figure is well-documented \citep{elder1993, pomerantz1977, ringach1996, marini2016}.
  \item[Continuity]to study the integration of individual elements across the visual field. When elements are oriented such that they would form a continuous object, humans are able to group the elements together to form a whole figure \citep{kwon2016, kovacs1993, field1993}.
  \item[Proximity]to study the cue of intra-element distance for grouping. When multiple elements in a figure are similar, humans can perceive many similar elements as belonging together with those that are closer in proximity \citep{kubovy2008, quinlan1998}.
  \item[Gradient Occlusion]to study the ability to handle full or partial occlusion of an element without a constant color value. The basis of objecthood in humans does not depend crucially on the exact color of the elements, and long-range integration and inference about objects and how they change under occlusion is possible \citep{kellman1983, spelke1990}.
  %\textcolor{red}{I think you once mentioned a horse-tree example. I think it is nice to mention it here to let the reader beware of the significant implications of the dataset.} ben: this is a different thing, this is just about occlusion, that one is about non-realistic extrapolation of a figure. although related, not the same thing. i edited the sentence above to clarify
  \item[Illusory Occlusion] to study grouping across the visual field while being occluded by an illusory object. Regular textures can be grouped together, and foreground objects that break the pattern can be separated from the complex textural background. A classic example is the Dalmatian Dog illusion by R. C. James, of which our Illusory Occlusion dataset is a simplification.
\end{description}

Together, the GG datasets comprise tests of a substantial number of the most major Gestalt laws of perceptual organization (for reviews, see \citet{wagemans2012a, wagemans2012b, jakel2016, todorovic2008, kanizsa1979}), including figure-ground segregation, proximity, continuity, closure, and the general notion of Good Gestalt (the tendency of the perceptual system to organize objects according to the aforementioned laws). 

\subsection{Evaluation metrics and baseline models}
\label{method:eval}
\textbf{Metrics.} We evaluate all models using the Adjusted Rand Index (ARI) metric \citep{hubert1985}, which measures the above-chance similarity between two clusterings: the target segmentation mask, and the predicted segmentation mask.
%Typically, the ARI-BG (ARI minus background) metric, where the background is removed from the evaluation, is used as a metric for judging how well a model segments objects from other objects.
%The reason behind this is that the ARI-BG is often superior in judging fine-grained details of object-specific segmentation capability, as compared to the ARI metric \textcolor{red}{(no need to mention this?)}.
Here we choose to judge our models based on the ARI metric (instead of ARI minus background) because the point of the GG datasets (and of the study of Gestalt in general) is that of object-background separation, not of fine-grained object-specific details.
We verified that in the GG datasets, an all-background assignment yields an ARI of $\approx0$, i.e. no better than random assignment.
In addition, we plot randomly drawn examples of segmentation masks in Figure \ref{fig:results}, and all segmentation masks for our highest-performing models in Appendix \ref{appendix:additional_outputs}, Figures \ref{fig:VAE_outputs_Kanizsa} - \ref{fig:AE_outputs_IllusoryOcclusion}.

\textbf{Models and control experiments.} 
We trained $5$ Autoencoder (AE) and Variational Autoencoder (VAE) models with MSE and GECO \citep{rezende2018} losses, respectively. 
%It is important to note that our goal in this work is not to propose a state of the art segmentation algorithm, but rather to understand the role of neuronal noise and its application to segmentation and grouping.
In addition, we computed two control experiments, one for each model variant (Noisy AE/VAE Reconstruction) with the goal of verifying whether adding noise at the latent layer is important.
These control experiments compute a clustering on the AE/VAE reconstruction pixel values, adding noise to the reconstruction instead of the latent layer, keeping the rest of the model architecture and methodology the same (for more details, see Appendices \ref{appendix:implementation} and \ref{fig:model_recons}).
%All models were implemented in PyTorch, and the Autoencoder and VAE models were trained on a high-performance cluster with a V100 GPU.
%, while baselines were trained on a high-performance computer with an RTX 4090 GPU \textcolor{red}{(I guess it is not necessary to mention the computation platform. It feels like selling ourselves short (because we train different models on different platforms). The training is reproducible on all platforms anyway. I think, usually, when people reveal the training platform, they also mention the training time to boast about their models convenience in use, which is not the flavor of this paper)}. 
%All models were trained for an approximately equivalent GPU-time. 
%Implementation details and model architectures are shown in Appendix \ref{appendix:implementation}, and reconstructions are shown in Appendix \ref{fig:model_recons}.
We also evaluate two baseline models, Genesis \citep{engelcke2020genesis} and Genesis-v2 \citep{engelcke2022}, for which details are shown in Appendix \ref{app:control_experiments}.

%\textbf{Baselines.} We compute several different baselines.
%The first two of these are a simple clustering based on reconstruction pixel values.
%While it is trivial that for complex, real-life datasets input values are not perfectly informative of object categories, we compute these values to ensure that we have developed testing datasets where good model reconstruction performance is not sufficient to segment the image well, despite the simplicity of the datasets.
%Hence, we do a similar procedure to our proposed model setup, where we stack reconstructions and cluster them --- the difference being that instead of injecting noise to the latent layer, we inject it directly to the reconstruction of the model, and cluster these noisy reconstructions to form segmentation masks.
%The next two baselines we compute are the AIR [ref] and Genesys [ref] models. 
%These models ... \textbf{TODO}. 
%The final model we compare performance to is a state-of-the-art object discovery model, SlotAttention, that explicitly assigns slots to different objects and learns to ... \textbf{TODO}. 
%Details on implementation of all baseline models can be found in Appendix \ref{appendix:implementation}.
%\textcolor{red}{GENESIS-V2 is more like SOTA, SlotAttention is quite old.
%STEGO is also really really good.
%There was also some model we mentioned about, like SAM, which is easy to use.
%Maybe we should also mention that competing the performance is not the aim of this paper, but rather we would like to highlight a principal for segmentation.}

\section{Experimental Results}
\label{section:results}

\subsection{Quantitative Evaluation}
\begin{table*}[h]
\centering
\setlength{\tabcolsep}{5pt} % adjust to suit your needs
\renewcommand{\arraystretch}{1} % adjust to suit your needs
%\begin{tabular}{lcccccc}
\begin{tabular}{llllllll}
\hline
\\[-2ex]
    Model & Kanizsa & Closure & Continuity & Proximity & Gradient Occ. & Illusory Occ. & Mean\\
    \hline
    \\[-2ex]
    AE & 0.859 {\tiny \(\pm\) 0.008} & 0.766 {\tiny \(\pm\) 0.008} & 0.552 {\tiny \(\pm\) 0.008} & \textbf{0.996*} {\tiny \(\pm\) 0.001} & 0.926 {\tiny \(\pm\) 0.009} & \textbf{0.994*} {\tiny \(\pm\) 0.002} & \textbf{0.849}\\
    VAE & \textbf{0.871} {\tiny \(\pm\) 0.005} & \textbf{0.795} {\tiny \(\pm\) 0.009} & 0.593 {\tiny \(\pm\) 0.012} & 0.943 {\tiny \(\pm\) 0.010} & 0.918 {\tiny \(\pm\) 0.002} & 0.974 {\tiny \(\pm\) 0.003} & \textbf{0.849} \\
    \hline
    \\[-2ex]
    AE Rec. & 0.246 {\tiny \(\pm\) 0.006} & 0.064 {\tiny \(\pm\) 0.001} & 0.570 {\tiny \(\pm\) 0.006} & 0.141 {\tiny \(\pm\) 0.001} & \textbf{0.982} {\tiny \(\pm\) 0.000} & -0.023 {\tiny \(\pm\) 0.001} & 0.333\\
    VAE Rec. & 0.343 {\tiny \(\pm\) 0.004} & 0.084 {\tiny \(\pm\) 0.000} & \textbf{0.611} {\tiny \(\pm\) 0.005} & 0.144 {\tiny \(\pm\) 0.001} & 0.977 {\tiny \(\pm\) 0.001} & -0.024 {\tiny \(\pm\) 0.001} & 0.356\\
    Genesis & 0.346  & 0.755 & 0.394  & 0.879 & 0.933  & 0.294 & 0.600\\
    Genesis-v2 & 0.415  & 0.000 & 0.399  & 0.059 & 0.000  & 0.422 & 0.216\\
    \hline
\end{tabular}
\caption{Model results (ARI \(\pm\) standard error of the mean). AE (Autoencoder) and VAE (Variational Autoencoder) model scores reported for the best noise level and number of samples. AE Rec. and VAE Rec. stand for noisy reconstruction control experiments, and (see Section \ref{method:eval} for details). Bold indicates top-performing model. For the AE and the VAE, a star (*) indicates a statistically significant difference in model performance (Bonferroni corrected). Our method improves performance over the previously published top-performing model \citet{engelcke2020genesis} by $0.249\%$.}
\label{table1}
\end{table*}
%Both the Autoencoder and VAE models segmenting using LNS achieve better or equal performance than the baselines on all datasets (Table \ref{table1}).
Both the AE and VAE models achieve good quantitative performance in general perhaps with the exception of the Continuity dataset (Table \ref{table1}). 
Adding noise in latent space as opposed to image space unlocks a substantial improvement in segmentation performance, improving ARI by $0.506$ on average across all GG datasets.
As described in Section \ref{section:qualitative}, we do not see the somewhat lower quantitative score as a failure \textit{per se}, but rather as related to the way in which the models fundamentally segment using LNS.
%\textcolor{red}{(I do not think this is a failure, we should say something like the following: The low ARI score for the Continuity dataset is a result of the model always perceiving thicker circles than the segmentation masks, which is overly punished by ARI.
%Nonetheless, it is with out doubt that the model can perceive the circles.) }
Since the segmentation process seeks to vary the latent representation of the models, it introduces artifacts related to the model's representational space, such as small shifts of the stimulus in a given direction, or small changes to its size.
The ARI metric is punishing to even single-pixel shifts of the target object --- qualitatively, it is clear that even in cases where the model does not achieve an excellent ARI score, it captures the desired effect.
This is in contrast to the Genesis and Genesis-v2 models, which fail at capturing the desired effect even when they do well quantitatively (Figure \ref{app:control_experiments}).
A similar effect of minor performance deterioration with respect to the baseline of clustering a reconstruction can be observed in the Gradient Occlusion dataset, where good reconstruction quality of the model is congruent with a good segmentation performance under output clustering.
This is to be expected - the dataset by design is such that the pixel values are mostly informative of the correct segmentation label identities.

%\textcolor{red}{(An important thing you did not mention: AE tends to get the wrong cluster numbers, because it fails to capture the covarying parts sometimes, like in Proximity and SwissFlag. 
%Our segmentation pipeline informs the model of the right cluster number beforehand, that's why VAE and AE achieve close performance.
%Without the prior information about group numbers, AE will not be able to group correctly. 
%It is better to make this case with some UMAP plots. 
%The  UMAP array is provided to you already, and I have some pilot code for plotting each one of them in the figures folder.)}

To understand whether a structured latent prior could improve model segmentation capability under LNS, we conducted a two-sample two-tailed t-test to compare the AE and VAE ($\beta>1$) performance.
The test indicated that the AE and VAE performance statistically significantly differed only for the Proximity ($t=4.54$, $p<0.05$) and Illusory Occlusion datasets ($t=5.22$, $p<0.05$) after the Bonferroni correction, while no statistically significant difference was found for the other datasets.
Due to the additional loss term that places tension on the quality of the reconstruction, it is not surprising that the VAE would show small decreases in performance --- indeed, we conclude that encouraging a structured latent prior on the model does not in general improve the model's capability to segment stimuli.% (but it is possible that the clusters formed by the VAE are more aligned with the desired results; for a further qualitative look, see Appendix \ref{appendix:umap}).
\begin{figure*}[h]
\centering
\includegraphics[scale=0.41]{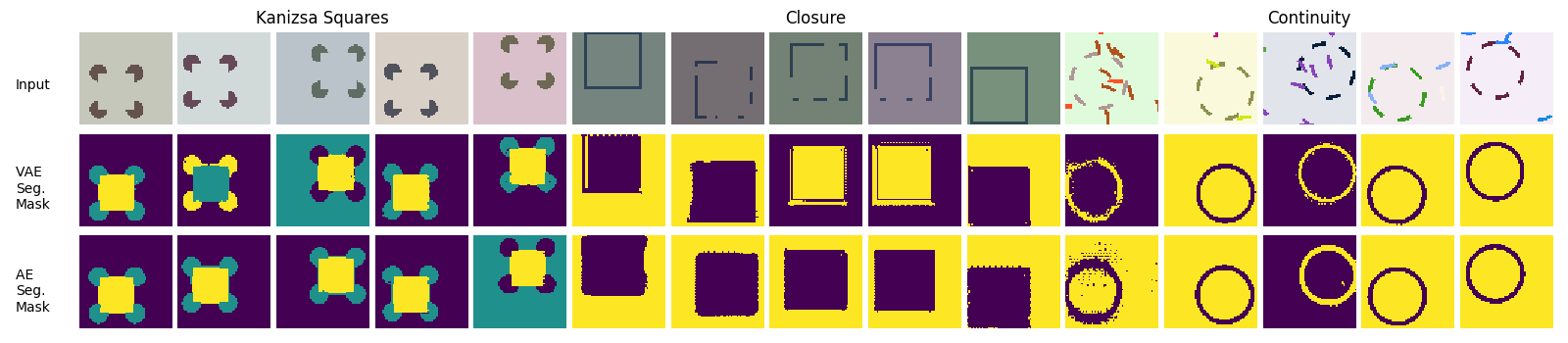}
\includegraphics[scale=0.41]{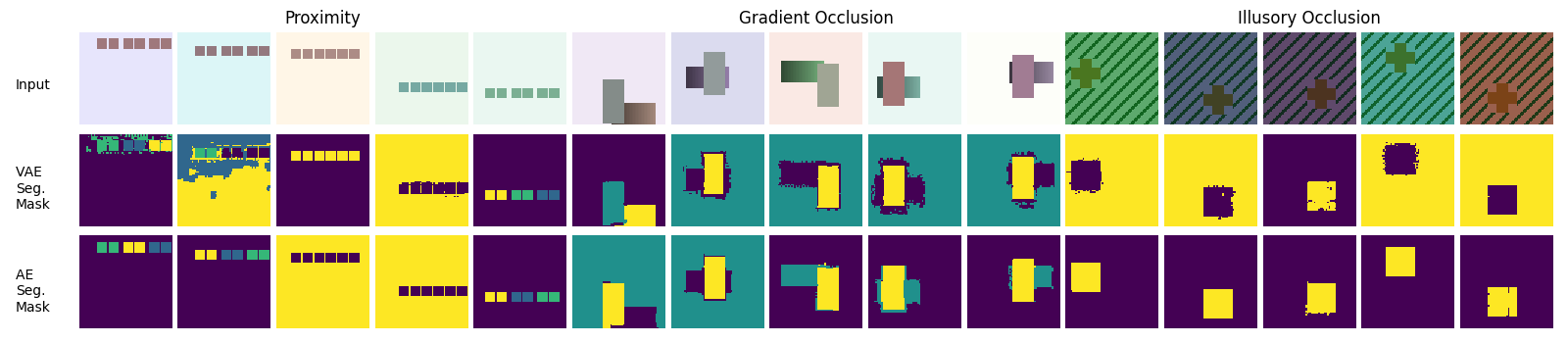}
\caption{\textbf{Model Output Segmentation Mask Examples.} The first row shows inputs, the second row shows VAE segmentation masks, and the third row shows AE segmentation masks. Randomly selected examples from the \textbf{Kanizsa Squares}, \textbf{Closure}, \textbf{Continuity}, \textbf{Proximity}, \textbf{Gradient Occlusion}, and \textbf{Illusory Occlusion} datasets are shown using the best model hyperparameters shown in Table \ref{table1}. Since the model segments by clustering noisy outputs, the specific color assignment to different object identities is arbitrary (for example, whether the model assigns the identity represented by yellow as the background, or purple, is not a meaningful distinction).}
\label{fig:results}
\end{figure*}
\subsection{Qualitative Evaluation}
\label{section:qualitative}

To understand how the models perform qualitatively, we plot a random selection of examples in Figure \ref{fig:results}, and all model outputs of the shown seed for all datasets in Appendix \ref{appendix:additional_outputs}.
Important are two things: first, while model outputs are consistently good and generally reflect appropriate grouping of the correct object elements in the correct groups, segmenting by noise causes noisy artifacts in some cases (e.g. Figure \ref{fig:results}: \textit{Continuity}, \textit{Gradient Occlusion}). 
Second, a qualitative inspection reveals how quantitative metrics fail to appropriately judge the goodness of model performance.
\begin{wrapfigure}{r}{0.3\textwidth}
\centering
\includegraphics[scale=0.5]{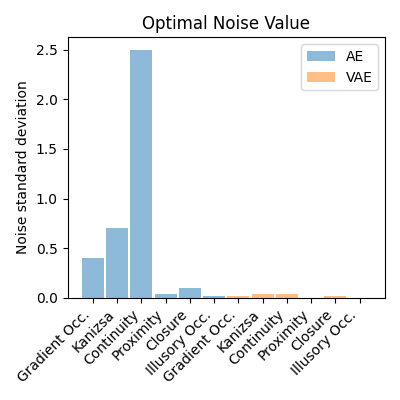}
\caption{\textbf{The optimal noise level for the AE and the VAE across datasets.} The AE performs best with a larger range of noise values, while the VAE consistently prefers low noise values.}
\label{fig:grouped_noise_values_colored_small_gap}
\end{wrapfigure}
Since segmentation is done by inference over a noisy set of outputs, and the noise affects the representation of those outputs, the outputs in some datasets (like \textit{Continuity} and \textit{Closure}) are often shifted or misshaped by a small number of pixels, resulting in a disproportionately large drop in quantitative performance, while qualitative performance remains relatively unchanged.

To further understand how elements are grouped in our model, we conducted two additional qualitative analyses: (1) Using UMAP visualization, we inspected model clustering and found that while the AE does not find the correct object binding, the VAE does (Appendix \ref{appendix:clustering}). (2) To understand how well the segmentation method generalizes, we qualitatively evaluated the models on the CelebA dataset \citep{liu2015} and found that while the AE generally fails to find a semantically meaningful segmentation mask (and instead opts for more basic image-level features), the VAE often finds a semantically meaningful segmentation of face-hair-background (Appendix \ref{appendix:celeba}).

\subsection{Noise sensitivity}

To understand how practically feasible LNS is for a system constrained by compute time or noise magnitude, we analyzed how the amount of noise added in latent space would affect the model’s segmentation performance.
While enforcing a prior distribution on the model's latent space (as is the case in the VAE) had no benefit in model performance (and perhaps a slight deterioration), it may have had a minor benefit in the across-task best-performing noise consistency.
We tested the same models across different levels of noise used for segmentation and found that enforcing the prior $\mathcal{N}(0, \mathbf{I})$ caused the model to consistently perform best at very small levels of noise, while the AE optimal level of noise varied across datasets (Figure \ref{fig:grouped_noise_values_colored_small_gap}).
This is intuitively not surprising, since the VAE enforces a specific activity magnitude for uninformative units in its prior ($\mathcal{N}(0, \mathbf{I})$), while the AE does not --- as such, the AE has no reason to prefer a specific coding magnitude in any given task.
%\textcolor{red}{(I think this explanation is unnecessary. It is not substantiated enough as well.
%The reason why there exists irrelevant units in VAE is due to the prior imposed on them during training. 
%Without the constraint, it could be the case that all the units in AE are relevant ones.
%Even if the argument here is true: AE is excessively affected by irrelevant units, the argument cannot explain why, in the presence of larger noise, the performance of AE is usually better than VAE. 
%My explanation is: VAE is doing local PCA, thus short-ranged perturbation can reveal the data distribution more accurately, while AE is doing PCA in a more global scale, so long-ranged perturbation reviews long-ranged PC directions. 
%It is just how different models represent things.
%Nothing more.)}.
To quantify the result, we performed an F-test for the equality of variances, and found that the level of noise that yielded the best results had higher variance for the AE than the VAE ($F=2994.56, p=1.11e^{-8}$).

The result suggests that enforcing a prior distribution on the representation may have an advantage in stabilizing the amount of noise needed to best segment stimuli across a variety of tasks.
That being said, the absolute performance of the AE model did not substantially decrease as a function of noise (Figure \ref{fig:quantitative} \textbf{left}), which means that it is possible that even the AE could yield meaningful segmentation results using a consistently low magnitude of noise.

\subsection{Number of time steps needed}
\begin{figure*}
\centering
\includegraphics[scale=0.135]{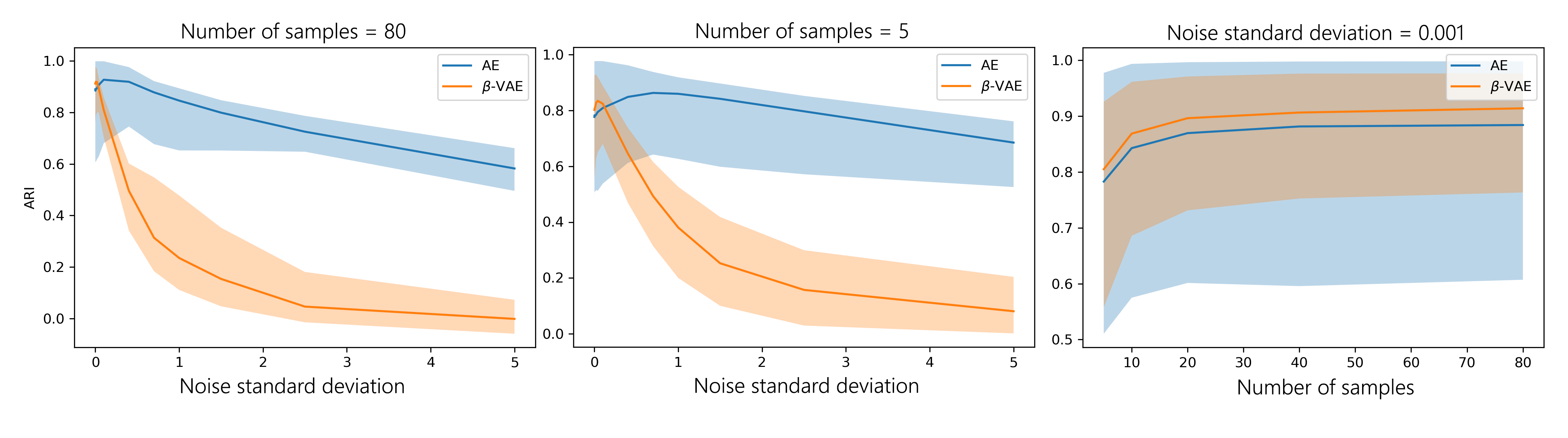}
\caption{\textbf{Sensitivity to Noise Magnitude and Time Steps.} \textbf{Left:} AE and VAE segmentation performance as a function of noise standard deviation, evaluated with $80$ time steps. \textbf{Middle:} AE and VAE segmentation performance as a function of noise standard deviation, evaluated with $5$ time steps. \textbf{Right:} AE and VAE segmentation performance as a function of the number of samples for a fixed noise $\sigma=0.001$. Note the change of $y$-axis scale. Shaded areas show 75th and 25th percentiles over all evaluated segmentation masks.}
\label{fig:quantitative}
\end{figure*}

We evaluated model performance as a function of the number of samples collected from the model that were used for segmentation (Figure \ref{fig:quantitative} \textbf{right}).
We found that for an appropriate level of noise used in segmentation, both the AE and the VAE models' performance asymptotes quickly, with as few as a handful of time steps. 
Reducing the number of time steps used from $80$ (Figure \ref{fig:quantitative} \textbf{left}) down to as few as $5$ barely reduces segmentation accuracy (Figure \ref{fig:quantitative} \textbf{middle}).

\section{Conclusion and Discussion}
We present Latent Noise Segmentation (LNS) that allows us to turn a deep neural network that is trained on a generic image reconstruction task into a segmentation algorithm without training any additional parameters, using only independent neural noise. 
The intuition behind why independent neural noise works for segmentation is that independent noise affects neurons that are selective to a presented stimulus more than those that are not.
%That neural noise works comes from the intuition that biological neural systems must cope with independent noise, and that in such a neural system, this noise affects neurons that are selective to a presented stimulus more than those that are not.
This idea is not entirely new --- noise has been shown to theoretically be able to \textit{enhance} the signal of a stimulus when the signal is small \mbox{\citep{benzi1981, kitajo2003, buchin2016}}; however, its application to segmentation and perceptual grouping, to the best of our knowledge, is novel. 

Importantly, our goal is not to propose a state of the art segmentation algorithm (though our model beats other tested baseline models), but rather to understand the role of neural noise and how it can be applied to achieve segmentation and perceptual grouping.
We implement LNS and find that it is possible to make pre-trained Autoencoders (AE) and Variational Autoencoders (VAE) segment images. 
We present the Good Gestalt (GG) datasets — a collection of datasets intended to investigate perceptual organization in humans — and show that LNS is able to explain the formation of ‘Good Gestalt‘ in a neural network model. 
Our findings give a unified high-level account of the result of many perceptual grouping phenomena in humans: 1) learning encodes an approximation of the training dataset distribution, and 2) neural noise reveals the learned representation in out-of-distribution samples, resulting in certain hallmarks of perception, like effects of illusory contours, closure, continuity, proximity and occlusion.
Establishing this link between training data (Appendix \ref{app: gg dataset details}), neural noise, and human perception is an important and timely step towards better models of visual representation \citep{bowers2022}. 

Interestingly, model learning rule plays a role in segmentation. 
We show that a very small amount of independent noise is optimal for segmentation in the case of the VAE, but not for the AE. 
Similarly, the VAE finds the correct number of objects in the image (Appendix \ref{appendix:clustering}) even when the AE does not.
This deviation between the VAE and the AE is interesting, as the VAE has connections with predictive coding \citep{joseph2022, boutin2020} and the free energy principle \citep{friston2010, mazzaglia2022}, and has been shown empirical support as a coding principle in primate cortex \citep{higgins2021}.
With both learning rules, segmentation performance asymptotes quickly, suggesting that the time scale of the segmentation in our models is practically feasible for constrained systems. 
%Together, our results show that it is possible to infer object identities even in ambiguous contexts using the model architecture itself, without having to train additional model parameters specifically for the task of segmentation.

These results suggest a number of interesting questions for future research. 
First, our focus here has been on revealing a potential mechanism for perceptual grouping and segmentation. 
While we have demonstrated that the approach partially scales to the case of face segmentation (Appendix \ref{appendix:celeba}), a relevant question is whether it is possible to scale the approach to improve unsupervised segmentation state-of-the-art \citep{bear2023, engelcke2022}.
Second, while we have targeted a large collection of most relevant grouping principles, there are many that we have not tested here. 
As we have shown for the datasets we have tested, we speculate that most or all phenomena in Gestalt perception can be reproduced as a result of a combination of past experience (learning) and noise in the perceptual system. 
Of interest is also understanding how LNS interacts with perception under a motion stream, which we have not considered here. 
Finally, while we do not make any claims about the primate visual system, our results suggest a potential benefit of independent noise in the visual system. 
Understanding how noise interacts with perception in primate cortex is an active area of research \citep{destexhe2012, miller2002, stein2005, guo2018, mcdonnell2011}, and we hope our results stimulate discussion about the potential benefits of noise for perceptual systems in general.

\input{code_and_data}
\section*{Impact Statement}
The paper presents work whose goal is to advance the field of Machine Learning. 
Our work makes improvements in unsupervised segmentation, and our work could contribute to potential societal consequences in several ways. 
Our work enables the use of neural noise for segmentation without labels, which could be used for harmful applications such as misuse of face segmentation. 
Since our methodology is unsupervised, its performance depends on the underlying trained model on which it is applied, and as such our methodology could make mistakes or exhibit biases against underrepresented groups. 
Furthermore, the addition of noise in the network could cause further mistakes by the model. 
In total, mitigating these risks by de-biasing underlying networks used for segmentation, and by verifying outputs is important in outcome-critical applications.

\section*{Acknowledgments}
We thank Jirko Rubruck, Martin Schrimpf, and Badr AlKhamissi for helpful discussions, and Marc Repnow for technical support. MHH and BL were supported by a grant from the Swiss National Science Foundation (N.320030\_176153; “Basics of visual processing: from elements to figures”).

\bibliography{iclr2024_conference}
\bibliographystyle{icml2024}

\newpage
\newpage
\appendix
\onecolumn
\section{Appendix}

\input{proof}

\newpage
\input{celeba.tex}

\newpage
\input{control_experiments.tex}

\newpage 
\input{clustering_algorithm.tex}

\newpage
\input{implementation_details.tex}

\newpage 
\input{umap.tex}

\newpage
\input{additional_outputs.tex}

\end{document}

%% file: math_commands.tex
\usepackage{amsmath}
\usepackage{bm}

\def\eqref#1{equation~\ref{#1}}
\def\1{\bm{1}}

\DeclareMathAlphabet{\mathsfit}{\encodingdefault}{\sfdefault}{m}{sl}
\SetMathAlphabet{\mathsfit}{bold}{\encodingdefault}{\sfdefault}{bx}{n}

%% file: code_and_data.tex
\section*{Reproducibility Statement}
Mathematical details about our results are included in Appendix \ref{appendix:proof}.
Full details on our GG datasets are included in Appendix \ref{app: gg dataset details}, and full details on model architecture and training are included in Appendix \ref{appendix:implementation}.
Code and data are public in our git repository: \hyperlink{https://github.com/ZhengqingUUU/LatentNoiseSegmentation}{github.com/ZhengqingUUU/LatentNoiseSegmentation}.

%% file: proof.tex
\subsection{Object separability}
\label{appendix:proof}
The Latent Noise Segmentation approach proposed in this work is capable of separating pixels into different objects regardless of their actual values in the input.
This fact is a necessary result yielded by a Variational Autoencoder (VAE) pursuing its optimization objective.
Here, we formalize the mathematical intuition behind why this is the case.

\subsubsection{Background}
\label{sec: background}

\cite{VAE_pursues_pca,demystifying_inductive_biases} showed that VAE-based Deep Neural Networks (DNNs) learn to represent local Principal Component (PC) axes.
The training process of a VAE aims to minimize the reconstruction loss, part of which can be re-assembled as the stochastic reconstruction loss: 

\begin{equation}
\hat{L}_{\mathrm{rec}}\left(\mathbf{x}^{i}\right)=\mathop{{}\mathbb{E}}\limits_{\mathbf{z}^i\sim \mathcal{N}(\boldsymbol{\mu}(\mathbf{x}^i), \operatorname{diag}(\boldsymbol{\sigma}^2(\mathbf{x}^i)))}\Vert\operatorname{Dec}_{\theta}\left(\boldsymbol{\mu}\left(\mathbf{x}^{i}\right)\right)-\operatorname{Dec}_{\theta}\left(\mathbf{z}^{i}\right)\Vert^{2},
\label{stochastic reconstruction loss}
\end{equation}
where the superscript $i$ denotes the index of the training samples.
% The expectation in Eq \eqref{stochastic reconstruction loss} is by taking the random variable, which is the $\operatorname{Enc}_{\varphi}(\mathbf{x}^i)$ over $\mathcal{N}(\boldsymbol{\mu}(\mathbf{x}^i), \operatorname{diag}(\boldsymbol{\sigma}^2(\mathbf{x}^i)))$.

We further define a Jacobian matrix to denote the decoded output's derivative with respect to the latent variables:
\begin{equation}
J_{i}=\left.\frac{\partial \operatorname{Dec}_{\theta}\left(\mathbf{z}\right)}{\partial \mathbf{z}}\right|_{\mathbf{z} = \boldsymbol{\mu}(\mathbf{x}^i)}\in \mathbb{R}^{D\times d},\label{eq: the Jacobian}
\end{equation}
where $D = img_x \times img_y\times C$ is the dimension of the images, and $d$ is the number of latent nodes.  
\cite{VAE_pursues_pca} proved that optimizing \eqref{stochastic reconstruction loss} will promote pairwise orthogonality on the columns of \eqref{eq: the Jacobian}, which means that the latent variables locally encode orthogonal features in the image space in $\mathbb{R}^D$. 
Furthermore, \citet{demystifying_inductive_biases} analyzed what those pairwise orthogonal directions are. 
With experiments, they empirically showed that those directions correspond to local PC directions.
More concretely, those directions are the PC directions yielded by a Principal Component Analysis (PCA) performed on a subset of all the training samples that lie close to each other in the data space.

Built on the theoretical framework of \citet{VAE_pursues_pca} and  \citet{demystifying_inductive_biases}, we are able to link these results to our context, and provide intuition for why the VAE is able to separate the object and the background.

One relevant technical detail is that the above statement regarding the latent nodes pursuing local local PC directions is derived from a VAE structure with a standardized objective as formulated by \citet{higgins2017}.
However, it is easy to check the proof from \citet{VAE_pursues_pca} to see that the same conclusion applies to our case, where the GECO \citep{rezende2018} loss is used. 

Furthermore, although the above-mentioned argument from \cite{VAE_pursues_pca, demystifying_inductive_biases} is for the Jacobians (\eqref{eq: the Jacobian}) calculated at training samples, we posit that the same conclusion holds for the testing samples.
After all, in our setting, the training samples and the testing samples are close to each other. 
Henceforth, $i$ will be used to denote the index of testing samples.

\subsubsection{Mathematical Intuition Underpinning Gestalt Perception} 

In the following, we will use $p_1$ to denote a pixel belonging to one object of the image, and $p_2$ to denote a pixel belonging to a different object of the image.

In our segmentation algorithm, we first repeatedly apply noise to the latent units and observe the resultant changes in output pixels. 
We denote the added noise, at the time step of $n$, to be $\boldsymbol{\xi}_n = (\xi_{n,j})_{j\in\{1,2,\cdots, d\}}^\intercal\in\mathbb{R}^d$, where the second subscript $j$ denotes the indices of the latent units to which noise is added.

Based on multivariate Taylor expansion, the reconstruction yielded by the latent affected by the noise can be written as:

\begin{equation}
\Tilde{\mathbf{x}}_n^i=\operatorname{Dec}\left(\boldsymbol{\mu}\left(\mathbf{x}^{i}\right)+\boldsymbol{\xi}_{n}\right)=\operatorname{Dec}\left(\boldsymbol{\mu}\left(\mathbf{x}^{i}\right)\right)+\left.\sum_{j=1}^{d}\xi_{n, j} \frac{\partial \operatorname{Dec}\left(\mathbf{z}\right)}{\partial z_j} \right|_{\mathbf{z} = \boldsymbol{\mu}(\mathbf{x}^i)}+o\left(\boldsymbol{\xi}_{n}\right),
\end{equation}
in which the superscript $i$ of $\mathbf{x}$ denotes the index of the input image, the subscript $j$ of $z$ denotes the latent index.
$\left.\frac{\partial \operatorname{Dec}\left(\mathbf{z}\right)}{\partial z_j} \right|_{\mathbf{z} = \boldsymbol{\mu}(\mathbf{x}^i)}$ is the $j^{th}$ column of the Jacobian in \eqref{eq: the Jacobian}.
$o\left(\boldsymbol{\xi}_{n}\right)$\ denotes the residual terms of the Taylor expansion.

In our segmentation algorithm, we subtract the reconstruction affected by one set of noise from the reconstruction affected by the previous set of noise. 
If we use $\boldsymbol{\delta}_{n} = (\delta_{n,j})_{j\in\{1,2,\cdots,d\}}^\intercal\in 
\mathbb{R}^d$ to represent $(\boldsymbol{\xi}_{n+1} - \boldsymbol{\xi}_{n})$, then we can write the difference between the two consecutive noisy reconstructions as:
\begin{equation}
\Delta\Tilde{ \mathbf{x}}_n^i=
\Tilde{ \mathbf{x}}_{n+1}^i - \Tilde{ \mathbf{x}}_{n}^i=\left.\sum_{j=1}^{d}\delta_{n, j} \frac{\partial \operatorname{Dec}\left(\mathbf{z}\right)}{\partial z_j} \right|_{\mathbf{z} = \boldsymbol{\mu}(\mathbf{x}^i)}\in \mathbb{R}^D,\label{output changes}
\end{equation}
where we omit the $o(\cdot)$ term, as they are negligible when the perturbing noise is sufficiently small.

Furthermore, we use $\Delta\Tilde{ \mathbf{x}}^i_{n,c,p_1}\in\mathbb{R}$ to denote the value of the component of $\Delta\Tilde{ \mathbf{x}}_n^i$ corresponding to the value of the pixel $p_1$ in the channel $c$ and  $\Delta\Tilde{ \mathbf{x}}^i_{n,c,p_2}\in\mathbb{R}$ to denote the value of the component of $\Delta\Tilde{ \mathbf{x}}_n^i$ corresponding to the value of the pixel $p_2$ in the channel $c$.

Our visual segmentation pipeline separates  pixel $p_1$ from pixel $p_2$ by separating the vector $\Delta\Tilde{ \mathbf{X}}_{p_1}^i = \left(\Delta\Tilde{ \mathbf{x}}^i_{n,c,p_1}\right)^\intercal_{(n,c) \in\{ 1,2,\cdots,N-1\}\times\{1,\cdots, C\}}\in\mathbb{R}^{(N-1)\times C}$ from the vector $\Delta\Tilde{ \mathbf{X}}_{p_2}^i = \left(\Delta\Tilde{ \mathbf{x}}^i_{n,c,p_2}\right)^\intercal_{(n,c) \in\{ 1,2,\cdots,N-1\}\times\{1,\cdots, C\}}\in\mathbb{R}^{(N-1)\times C}$. 

More concretely, as indicated in Appendix ~\ref{sec: clustering detail}, the separation between the two vectors are based on the angles they point toward in the $\left[(N-1)\times C\right]$-dimensional space.
In other words, the case in which the two pixels cannot be separated with our pipeline can only be caused by the case that no matter how we sample our noise, the following condition holds with probability $1$. 
\begin{description}
    \item[Condition for Inseparability.] There exist some constant $r_{p_1, p_2}> 0$ such that, for the given pixel pair $(p_1, p_2)$ and any channel $c$ and any time step $n$: 
$
\Delta\Tilde{ \mathbf{x}}^i_{n,c,p_1} = r_{p_1,p_2} \Delta\Tilde{ \mathbf{x}}^i_{n,c,p_2}
$.
\end{description}

The equation in the condition for inseparability can be expanded as: 
\begin{equation}
\left.\sum_{j=1}^{d}\delta_{n, j} \frac{\partial \operatorname{Dec}\left(\mathbf{z}\right)_{c,p_1}}{\partial z_j} \right|_{\mathbf{z} = \boldsymbol{\mu}(\mathbf{x}^i)} = r_{p_1,p_2} \left.\sum_{j=1}^{d}\delta_{n, j} \frac{\partial \operatorname{Dec}\left(\mathbf{z}\right)_{c,p_2}}{\partial z_j} \right|_{\mathbf{z} = \boldsymbol{\mu}(\mathbf{x}^i)}\label{eq: expanded equality}
\end{equation}
The fact that condition for inseparability holding with probability $1$ implies inseparability is intuitive, and we also provide a proof for it in \ref{stringent reasoning for separation probability}.

In the following, we will argue why the condition for inseparability is unlikely to hold with probability $1$ given the background in Section~\ref{sec: background}.

For the condition of inseparability to hold, the Jacobian (in \eqref{eq: the Jacobian}) and the noise must satisfy one of the following two cases.
 
The \textbf{first case} is when the Jacobian satisfies: 
 \begin{equation}
     \left.\frac{\partial \operatorname{Dec}\left(\mathbf{z}\right)_{c,p_1}}{\partial z_j} \right|_{\mathbf{z}= \boldsymbol{\mu}(\mathbf{x}^i)}  =  r_{p_1, p_2}    \left.\frac{\partial \operatorname{Dec}\left(\mathbf{z}\right)_{c,p_2}}{\partial z_j} \right|_{\mathbf{z}= \boldsymbol{\mu}(\mathbf{x}^i)}
     \label{eq: unrealistic local PC relationship}
 \end{equation}
 for some fixed $r_{p_1,p_2}$ for all $j\in\{1,2,\cdots,d\}$ and all channels $c$. 
 Here $\operatorname{Dec}\left(\mathbf{z}\right)_{c,p_2/p_2}\in \mathbb{R}$ is the component of the $D$-dimensional vector  $\operatorname{Dec}\left(\mathbf{z}\right)$ that corresponds to the $p_1$ or $p_2$ pixel in channel $c$.
In this case, the condition for inseparability trivially holds with probability $1$.
 
The \textbf{second case} is where \eqref{eq: unrealistic local PC relationship} does not hold for any fixed $r_{p_1,p_2}$ for some $j\in\{1,2,\cdots,d\}$ and some channel $c$.
In this case, the condition for inseparability might still hold.

In the following, we will explain why both cases cannot hold with probability $1$ when $p_1$ and $p_2$ are pixels belonging to different objects of the image, which means that they have independent pixel values in the training dataset (see Appendix~\ref{app: gg dataset details} for more details).

\subsubsection*{(1) Why is the first case unlikely to hold?}

The argument of the first case boils down to a characterization of the Jacobian in \eqref{eq: the Jacobian}.
A complete theoretical understanding of the Jacobian has yet to be established in the past literature to the best of our knowledge. 
However, from the conclusions derived from \citet{VAE_pursues_pca} and \citet{demystifying_inductive_biases}, which we briefly mentioned in Appendix~\ref{sec: background}, the first case is unlikely to happen as it intuitively goes against the behavior of PCA.

In more detail, as demonstrated in \cite{demystifying_inductive_biases}, the columns of the Jacobian (\eqref{eq: the Jacobian}) should point into directions in the $D$-dimensional space that are principal components extracted from a compact cluster of training images in the $D$-dimensional space.
In addition, \eqref{eq: unrealistic local PC relationship} implies that all the principal components characterize the pixel values (in each channel) of $p_1$ and $p_2$ to have a strict linear relationship with the same ratio $r_{p_1 p_2}$.
However, as introduced in Appendix~\ref{app: gg dataset details}, the training dataset is generated by assuming the pixel values of $p_1$ and $p_2$ (when they do not belong to the same object) are independent. 
Thus, it is unlikely to have all local PC directions satisfying \eqref{eq: unrealistic local PC relationship}.

\begin{figure}[ht]
\begin{subfigure}{.5\textwidth}
  \centering
  % include first image
  \includegraphics[width=.8\linewidth]{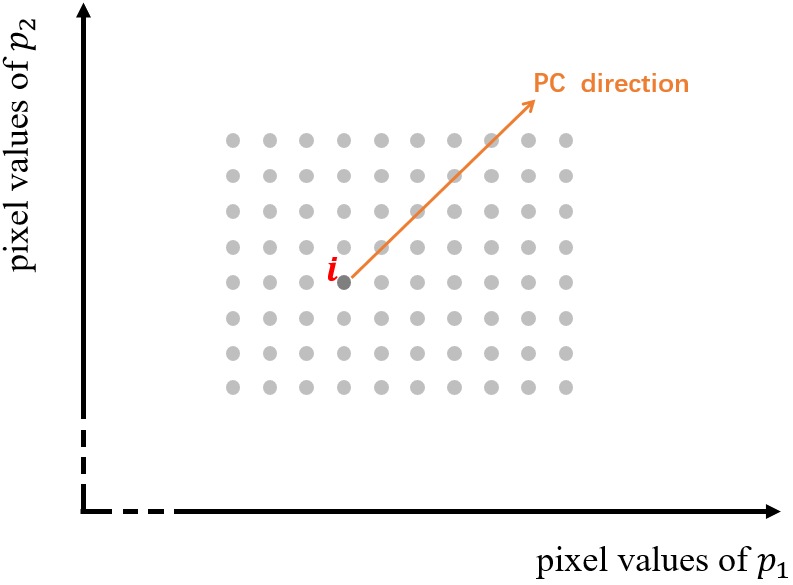}  \caption{$p_1$ and $p_2$ belong to different parts}
  \label{fig:sub-first}
\end{subfigure}
\begin{subfigure}{.5\textwidth}
  \centering
  % include second image
  \includegraphics[width=.8\linewidth]{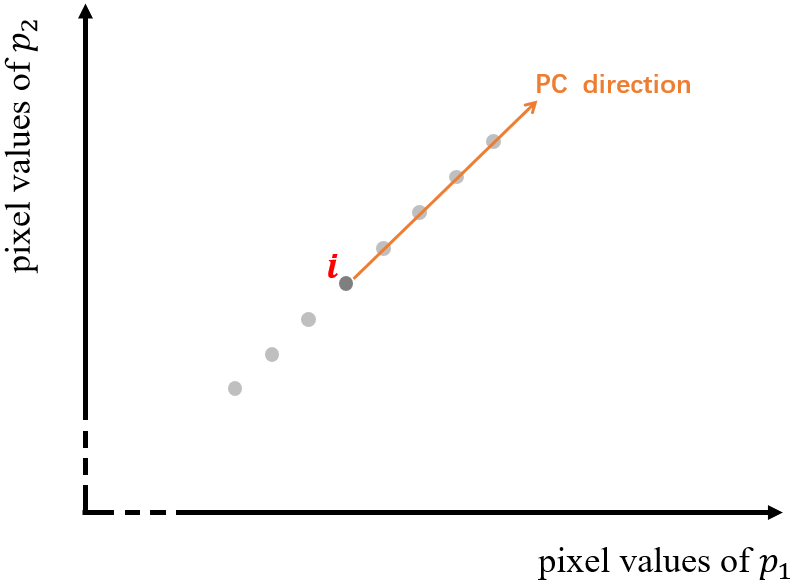}  
  
  \caption{$p_1$ and $p_2$ belong to the same part}
  \label{fig:sub-second}
\end{subfigure}
\caption{\textbf{(Left)} An example of a direction that all the local PC directions should follow to have $p_1$ and $p_2$ become inseparable. \textbf{(Right)}
An example of how the local PC directions might look like when they are projected into the pixel value space of two pixels of the same part of the images.
}
\label{fig: abnormal local PC direction}
\end{figure}

For clarity, we provide a diagram in the left panel of Figure~\ref{fig: abnormal local PC direction}. 
Here we project the entire training dataset (denoted by the gray dots) to a 2-dimensional space. 
Each dimension stands for the pixel value of $p_1$ or $p_2$.
The grey dots are arranged in a way to demonstrate the independence of the two pixel values.
The orange arrow denotes a potentially learned local PC direction, moving along which the variation of the two pixel values follows a strict linear ratio. 
Since there are data variances in other directions, it is highly unlikely that all the local PC directions learned from local data points only capture directions of such kind.

On the other hand, we can also develop intuition into why any two pixels belonging to the same object can be inseparable with the right panel of Figure~\ref{fig: abnormal local PC direction}.
If $p_1$ and $p_2$ are pixels belonging to the same object, the way the dataset is generated (discussed in Appendix \ref{appendix:implementation}) dictates that the training samples projected onto the pixel value space will only stretch in a direction where there exists a strict linear ratio between the values of the two objects.
In this case, the learned local PC directions are highly likely to point only in the direction on this $2$-dimensional plane, as there exists no data variance along other directions. 
Such learned local PC directions guarantee that the two pixels of the same object will not be separated in the segmentation process, for it implies the Jacobian satisfies the condition described in \eqref{eq: unrealistic local PC relationship} in the first case.

\subsubsection*{(2) Why can the second case not hold with probability $1$?}

To understand this case, we first rewrite \eqref{eq: expanded equality} as:
 
 \begin{equation}
\sum_{j=1}^{d}\left[ \left.\frac{\partial \operatorname{Dec}\left(\mathbf{z}\right)_{c,p_1}}{\partial z_j} \right|_{\mathbf{z} = \boldsymbol{\mu}(\mathbf{x}^i) } - r_{p_1,p_2}  \left.\frac{\partial \operatorname{Dec}\left(\mathbf{z}\right)_{c,p_2}}{\partial z_j} \right|_{\mathbf{z} = \boldsymbol{\mu}(\mathbf{x}^i) }\right] \delta_{n,j}
= \sum_{j = 1}^d C_{c,j}(r_{p_1,p_2}) \delta_{n,j}= 0, \label{eq: the hyperplane expression}
\end{equation}
in which $C_{c,j}(r_{p_1,p_2})$ is used to summarize the term in the square bracket $\left[\hspace{0.2cm}\cdot\hspace{0.2cm}\right]$.
As we are considering the second case, we know that $C_{c,j}(r_{p_1,p_2})\neq 0$ for at least one $j\in \{1,2, \cdots, d \}$ and one channel $c$.
Thus, in the second case, the condition for inseparability can be interpreted as:  $\boldsymbol{\delta}_{n} = (\delta_{n,1},\delta_{n,2},\cdots,\delta_{n,d})^\intercal\in 
\mathbb{R}^d$ lies in a hyperplane as described by \eqref{eq: the hyperplane expression}, and the hyperplane has less than $d$ dimensions.
Such an event cannot happen with probability $1$.
This is because $\boldsymbol{\delta}_{n}$ has a Gaussian distribution in the $d$-dimensional space, implying its probability mass on a hyperplane with lower dimensions is $0$.

\subsubsection{Why Probability $1$ Implies Inseparability}
\label{stringent reasoning for separation probability}

Here, we give a formal derivation of why the condition for inseparability must hold with probability $1$ to make $p_1$ and $p_2$ inseparable. 
To simplify notation, let us denote the event that \eqref{eq: expanded equality} holds for some time step $n$ some channel $c$ and some positive constant $r_{p_1, p_2}$ to be $\mathcal{A}_{n, c}(r_{p_1,p_2})$. 
It would require $\mathcal{A}_{n,c}(r_{p_1,p_2})$ to hold for all $n\in \{1,2,\cdots, N-1\}$ and all $c\in \{1,\cdots,C\}$ to make $p_1$ and $p_2$ inseparable, otherwise $\Delta\Tilde{\mathbf{X}}^i_{ p_1}$ and  $\Delta\Tilde{\mathbf{X}}^i_{ p_2}$, both belonging to $\mathbb{R}^{(N-1)\times C}$, do not point into the same direction, making the two pixels separable. 

Intuitively, if the condition for inseparability does not hold with probability $1$, then the events $\{\mathcal{A}_n(r_{p_1,p_2})\}_{n\in \{1,2,\cdots, N-1\}}$ are less and less likely to happen simultaneously as $N$ increases, meaning $p_2$ and $p_1$ are always separable given $N$ sufficiently large.
Here below, we characterize this intuition rigorously.

Note that $\mathcal{A}_{n,c}(r_{p_1,p_2})$ and $\mathcal{A}_{n+1,c}(r_{p_1,p_2})$ may be dependent on each other, as we can show that $\boldsymbol{\delta_{n}}$ and $\boldsymbol{\delta_{n+1}}$ are dependent of each other.
$\boldsymbol{\delta_{n}}$ involves the noise pair of $(\boldsymbol{\xi}_n,\boldsymbol{\xi}_{n+1})$ and $\boldsymbol{\delta_{n+1}}$ involves the noise pair of $(\boldsymbol{\xi}_{n+1},\boldsymbol{\xi}_{n+2})$.
Both are affected by $\boldsymbol{\xi}_{n+1}$ and are thus dependent (a more formal explanation of which will be given in the last part of this section). 
Nonetheless, we have that 
$\mathcal{A}_{n,c}(r_{p_1,p_2})$ and $\mathcal{A}_{n+2,c}(r_{p_1,p_2})$ are independent of each other, since the former concerns the noise pair of $(\boldsymbol{\xi}_n,\boldsymbol{\xi}_{n+1})$ and the latter concerns the noise pair of $(\boldsymbol{\xi}_{n+2},\boldsymbol{\xi}_{n+3})$, which are unaffected by each other.

Thus, suppose we arbitrarily choose a channel $c_0$:
\begin{subequations}
\begin{align}
&\mathbb{P}(\text {inseparability})\leq\mathbb{P}\left(\mathcal{A}_{1, c_0}(r_{p_1,p_2}),\mathcal{A}_{2, c_0}(r_{p_1,p_2}), \cdots, \mathcal{A}_{N-1, c_0}(r_{p_1,p_2})\right) \\
&\leq \mathbb{P}\left(\mathcal{A}_{1,c_0}(r_{p_1,p_2}),\mathcal{A}_{3,c_0}(r_{p_1,p_2}), \cdots, \mathcal{A}_{{2}\lceil\frac{N-1}{2}\rceil-1, c_0}(r_{p_1,p_2})\right) \\
&=\prod_{i=1}^{\lceil\frac{N-1}{2}\rceil} \mathbb{P}\left(\mathcal{A}_{2i-1,c}(r_{p_1,p_2})\right) \label{goes to zero}
\end{align}
\end{subequations}
The independence between $\mathcal{A}_{n,c_0}(r_{p_1,p_2})$ and $\mathcal{A}_{n+2,c_0}(r_{p_1,p_2})$ enables the fully factorized representation in  \eqref{goes to zero}. 
Note that every probability term in \eqref{goes to zero} has the same value, and if they are all smaller than $1$ (for all $r_{p_1, p_2}>0$), the product will decay to zero exponentially (for all $r_{p_1, p_2}>0$). 
Thus, we require \eqref{eq: expanded equality} to happen with the probability $1$ to achieve inseparability.

The last part of this section demonstrates why $\boldsymbol{\delta}_{n}$ and $\boldsymbol{\delta}_{n+1}$ are not independent.
If $\boldsymbol{\delta}_{n}$ and $\boldsymbol{\delta}_{n+1}$ are independent, then we must have that the variance of their sum should be equal to the sum of their variances. However,
\begin{equation}
\begin{aligned}
&\operatorname{Var}\left(\boldsymbol{\delta}_{n+1}+\boldsymbol{\delta}_{n}\right) =\operatorname{Var}\left(\boldsymbol{\xi}_{n+2}-\boldsymbol{\xi}_{n+1}+\boldsymbol{\xi}_{n+1}-\boldsymbol{\xi}_{n}\right) =\operatorname{Var}\left(\boldsymbol{\xi}_{n+2}-\boldsymbol{\xi}_{n}\right)=2 \boldsymbol{\sigma}_{\boldsymbol{\xi}}^{2} \\
& \neq 4 \boldsymbol{\sigma}_{\boldsymbol{\xi}}^{2} =\operatorname{Var}\left(\boldsymbol{\xi}_{n+2}-\boldsymbol{\xi}_{n+1}\right)+\operatorname{Var}\left(\boldsymbol{\xi}_{n+1}-\boldsymbol{\xi}_{n}\right) =\operatorname{Var}\left(\boldsymbol{\delta}_{n+1}\right)+\operatorname{Var}\left(\boldsymbol{\delta}_{n}\right), \label{dependency of nearby noise}
\end{aligned}
\end{equation}
resulting in a contradiction.
With slight abuse of notation, all the terms in \eqref{dependency of nearby noise} are $d$-dimensional vectors, and all the operations are component-wise.

\subsubsection{Discussion}

With the above, we demonstrated how the underlying local PCA enables VAE to segment the images as we desired. 
In our experiments, we also show that the Autoencoders can also manage to segment the images similarly, but that their preferred number of objects is different to the VAEs'. 
Indeed, qualitatively, the Autoencoder generalizes worse than the VAE (Appendix \ref{appendix:celeba}).
Previous works have drawn a link between PCA and Linear Autoencoders \citep{linear_ae_pca}.
However, a full theoretical account for the connection between PCA and the more complicated non-linear autoencoder to the best of our knowledge does not yet exist.
With the intuitions provided in the experimental observation in this paper, we are at a better place to formulate the connection, which we consider as a meaningful future direction. 
Furthermore, we have provided a mathematical link between perception in humans, representation in Deep Neural Networks, and semantic segmentation.
We hope our work encourages future work in forming rigorous links between these established fields.

\subsubsection{Applicability to Other Generative Algorithms}
\label{app:proof_discussion}
In this work, we have focused primarily on the Variational Autoencoder (VAE) in lieu of other generative models, such as Diffusion models \citep{sohl-dickstein2015, rombach2022} or Adversarial Generative \citep{goodfellow2014, creswell2018} models.
There are two primary reasons behind this choice: 1) The VAE formulation allows us to neatly follow the argumentation in \cite{demystifying_inductive_biases} and \cite{VAE_pursues_pca} to arrive at a more firm understanding of why Latent Noise Segmentation (LNS) works; 2) the mathematical and computational links between the VAE formulation and other biologically plausible coding and learning schemes such as predictive coding \citep{boutin2020, joseph2022} and the Free Energy Principle \citep{friston2010, mazzaglia2022} are clearer than for other generative algorithms to the best of our knowledge.

That being said, the fact that the Autoencoder works relatively well for segmentation using LNS is interesting, and raises the question of whether other model architectures could also perform well.
We suggest that future work could empirically evaluate LNS on other model classes, such as Latent Diffusion models \citep{rombach2022} to understand whether the mathematical intuition behind why LNS works also generalizes beyond the simple Autoencoder model class to any model class that learns meaningful representations about images.

It is important to also note that the noise in Latent Noise Segmentation (LNS) acts in a fundamentally different way than noise in other generative models, such as diffusion models. 
While diffusion models are trained to denoise data in a stepwise manner, noise in LNS is not a part of the training process, and is only added at inference to stochastically explore the neighborhood of the latent representation.
While both models are stochastic and add noise, noise serves a functionally different role in each: a key role in training in diffusion models; and an easy way to stochastically explore the local neighborhood of the input stimulus in LNS.

%% file: celeba.tex
\section{Generalization to More Naturalistic Stimuli}
\label{appendix:celeba}
To test the generalization and potential scalability of our models, we evaluated them qualitatively on the CelebA dataset \citep{liu2015}. 
The dataset consists of a large variety of faces in different contexts, lighting conditions, and backgrounds. 
We trained the model on the CelebA training dataset as provided, and tested on the first 100 samples of the testing dataset.

\begin{figure}[h]
\begin{center}
%\framebox[4.0in]{$\;$}
\includegraphics[scale=0.25]{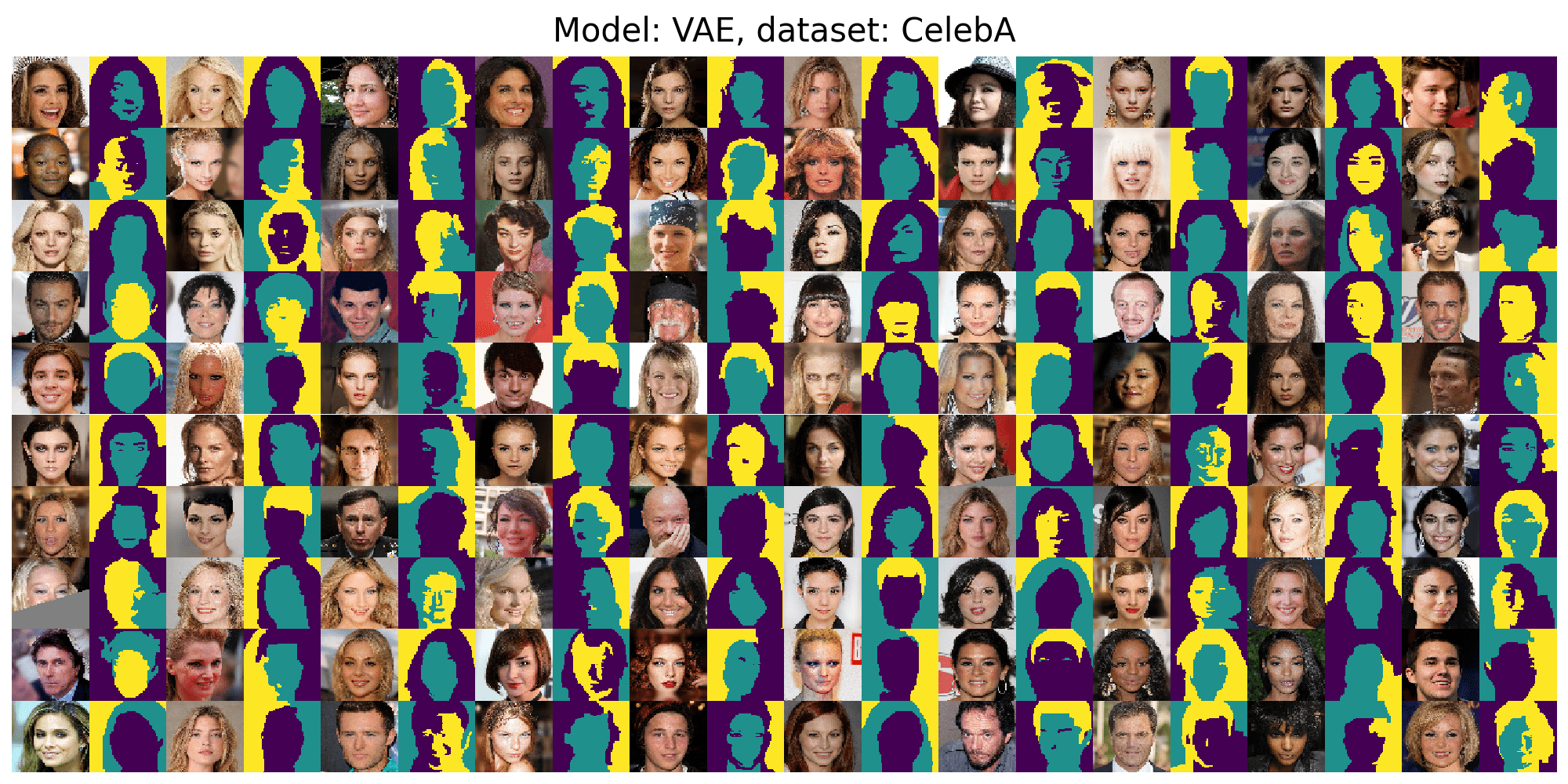}
\end{center}
\caption{\textbf{All tested Variational Autoencoder outputs for a single seed of the CelebA dataset.} Inputs and outputs are displayed as horizontal pairs, with the input being the left image, and the model output being the right image.}
\label{fig:VAE_outputs_CelebA}
\end{figure}

\begin{figure}[h]
\begin{center}
%\framebox[4.0in]{$\;$}
\includegraphics[scale=0.25]{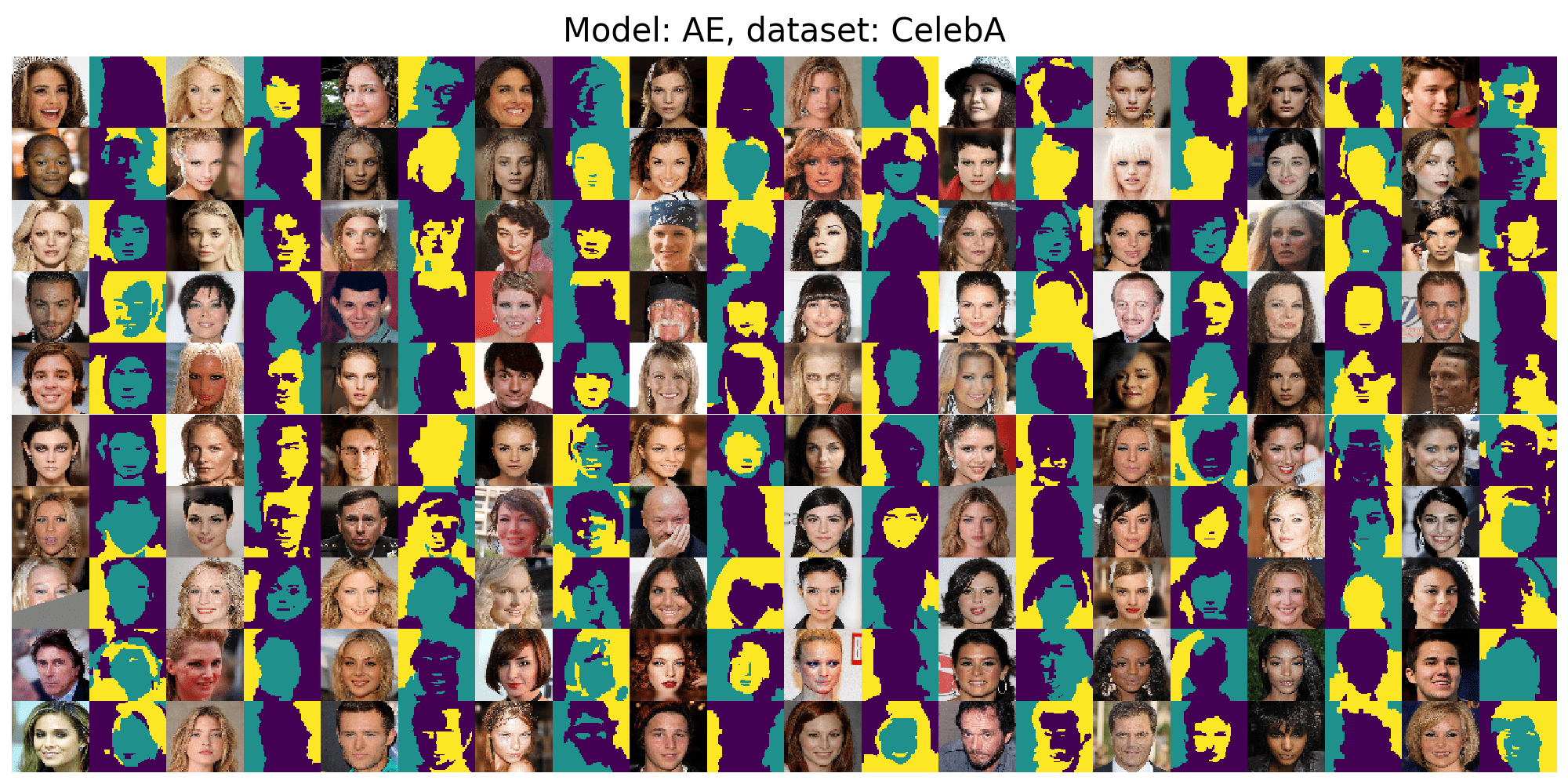}
\end{center}
\caption{\textbf{All tested Autoencoder outputs for a single seed of the CelebA dataset.} Inputs and outputs are displayed as horizontal pairs, with the input being the left image, and the model output being the right image.}
\label{fig:AE_outputs_CelebA}
\end{figure}

We found that the VAE model successfully finds a desired face-hair-background segmentation mask relatively often, when prompted on $3$ output clusters (Figure \ref{fig:VAE_outputs_CelebA}). 
In some cases the clustering is coherent but not semantically aligned with the desired output --- we suspect that in the CelebA dataset, lighting condition forms one of the strongest local PC components, and that this leads to a semantically less meaningful cluster on occasion.
This effect is seen more strongly in the case of the AE, which appears to primarily follow a lighting-related clustering (Figure \ref{fig:AE_outputs_CelebA}).

%% file: control_experiments.tex
\section{Control Experiments}
\label{app:control_experiments}
\subsection{Additional Control Experiment Results}
\begin{figure}[h]
\centering
\includegraphics[scale=0.35]{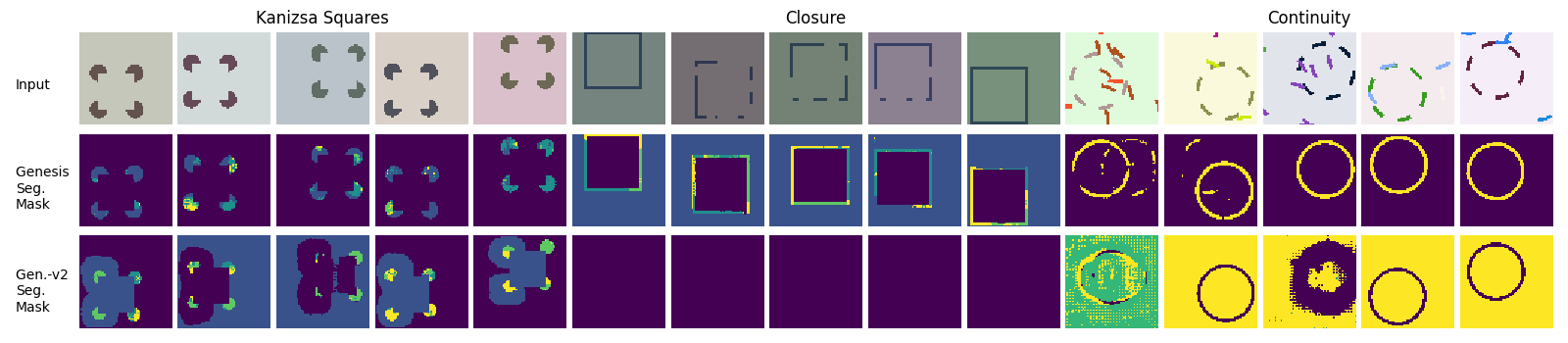}
\includegraphics[scale=0.35]{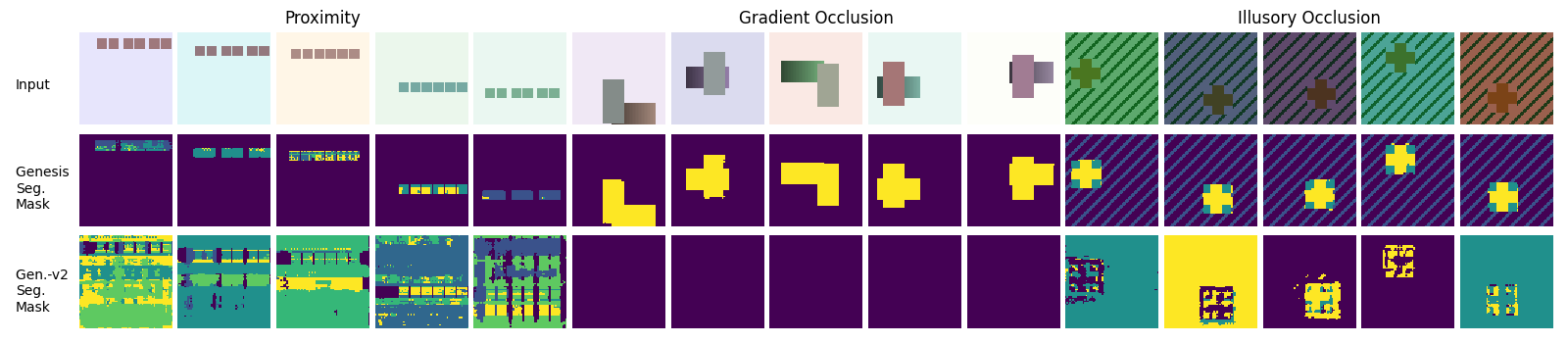}
\caption{\textbf{Control Model Segmentation Mask Examples.} The first row shows inputs, the second row shows Genesis segmentation masks, and the third row shows Genesis-v2 segmentation masks. Randomly selected examples from the \textbf{Kanizsa Squares}, \textbf{Closure}, \textbf{Continuity}, \textbf{Proximity}, \textbf{Gradient Occlusion}, and \textbf{Illusory Occlusion} datasets. The specific color assignment to different object identities is arbitrary (for example, whether the model assigns the identity represented by yellow as the background, or purple, is not a meaningful distinction).}
\label{fig:baseline_results}
\end{figure}

We tested two high-performing unsupervised object segmentation models from the literature: Genesis \citep{engelcke2020genesis} and Genesis-v2 \citep{engelcke2022}. 
We show quantitative results in Table \ref{table1}.
While Genesis performs quantitatively relatively well, a closer inspection of the actual segmentation performance of the model shows that with the exception of the Closure and Continuity datasets, it fails to find the correct Gestalt grouping of the stimuli \ref{fig:baseline_results}, often opting to follow the color value cues in the testing dataset.
The Genesis-v2 performs generally worse than Genesis, failing on all datasets both quantitatively and qualitatively with perhaps the exception of the Continuity dataset. 

\subsection{Implementation Details And Convergence}
\begin{figure}[h]
\centering
\includegraphics[scale=0.35]{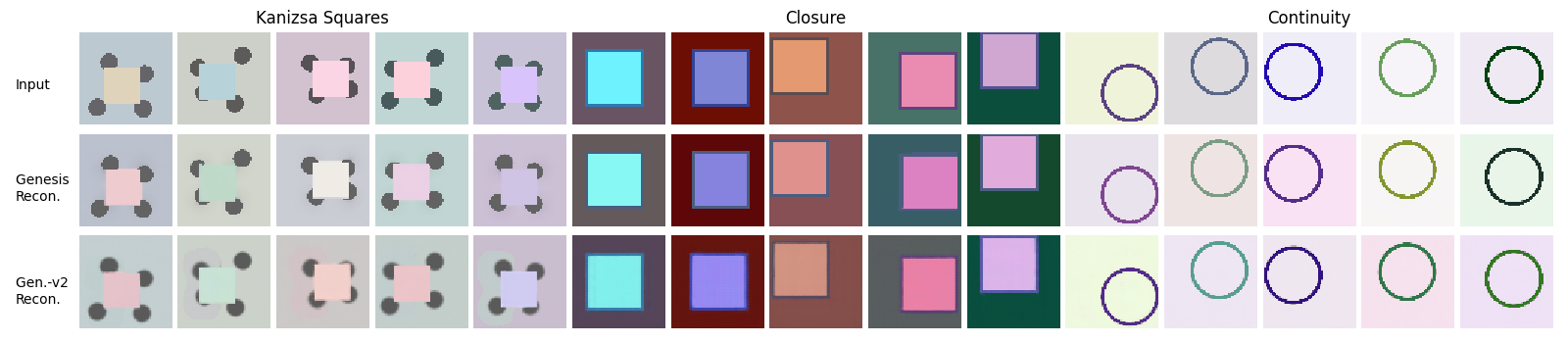}
\includegraphics[scale=0.35]{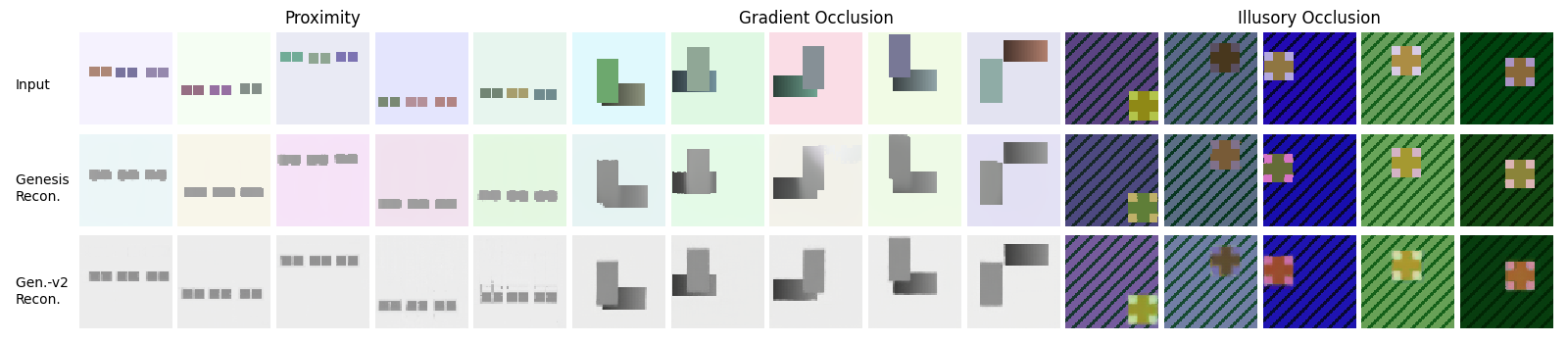}
\caption{\textbf{Control Model Reconstruction Examples.} The first row shows inputs, the second row shows Genesis reconstructions, and the third row shows Genesis-v2 reconstructions. Randomly selected examples from the \textbf{Kanizsa Squares}, \textbf{Closure}, \textbf{Continuity}, \textbf{Proximity}, \textbf{Gradient Occlusion}, and \textbf{Illusory Occlusion} validation datasets. The validation samples come from the same distribution as the training data, but the specific samples are never seen in training.}
\label{fig:baseline_recons}
\end{figure}

To ensure that the baseline failure is not due to an implementation error or a failure in convergence, we take two precautions:
1) We follow the official implementation of Genesis and Genesis-v2 \citep{engelcke2022, engelcke2020genesis, engelcke2020reconstruction} (GPL v3 license);
2) We verify convergence by visualizing model reconstructions on a separate validation set with samples that were not shown during training \ref{fig:baseline_recons}.
Furthermore, we follow the hyperparameter choices reported in the original papers ($5e^5$ training iterations with batch size $64$, $K=5$). Models were trained using an RTX 4090 GPU. Genesis-v2 trained for approximately 30 GPU-hours per dataset, while Genesis trained for approximately 20 GPU-hours per dataset, for a combined total of approximately 300 GPU-hours.
\subsection{Noisy AE/VAE Control Experiment Implementation Details}
The noise standard deviation applied to the reconstructions is $0.3235$, the mean of the best noise values for the AE and VAE. For clustering we used Agglomerative Clustering with the Euclidean metric and Ward linkage.

%% file: clustering_algorithm.tex
\section{Clustering}
\label{sec: clustering detail}
We used Agglomerative Clustering \citep{scikit-learn} with the Euclidean metric and Ward linkage.
We selected the desired number of clusters on an image-by-image basis. 
The desired number of clusters is the number of co-varying parts in the visual scene.
For example, for an image in the GG Proximity dataset, if it contains one group of $6$ squares, then the desired number of clusters is $2$, corresponding to the co-varying squares and the background; if it contains $3$ groups of $2$ squares, then the desired number of clusters is $4$, corresponding to the three groups of co-varying squares and the background.

Prior to applying the agglomerative clustering algorithm,  we normalize $\Delta \Tilde{\mathbf{X}}$ along its last dimension, so the vectors (with a size of $[\text{C}\times{(N-1)}]$) have the same norm.

\subsection{Comparison of Different Clustering Algorithms}
We conducted a control experiment to test whether the specific clustering we used is a critical part of why LNS works. 
We tested three clustering algorithms: Agglomerative Clustering (\citep{scikit-learn} with the Euclidean metric and Ward linkage; Agglomerative Clustering with the Euclidean metric and Complete linkage; and K-means clustering.
We find that while the choice of clustering algorithm can affect results (Ward linkage outperforms Complete linkage), the specific algorithm is not a critical part of LNS (K-means achieves similar or better performance than Agglomerative Clustering with the Ward linkage).
This result supports the conclusion that the key intuition behind LNS is that it allows the model to extract a meaningful clusterable representation, that can then be clustered using various clustering algorithms.
\begin{table}[h]
\centering
\setlength{\tabcolsep}{5pt} % adjust to suit your needs
\renewcommand{\arraystretch}{1} % adjust to suit your needs
%\begin{tabular}{lcccccc}
\begin{tabular}{lllllll}
\hline
\\[-2ex]
    Model & Kanizsa & Closure & Continuity & Proximity & Gradient Occ. & Illusory Occ. \\
    \hline
    \\[-2ex]
    AE+Agg & 0.859 {\tiny \(\pm\) 0.008} & 0.766 {\tiny \(\pm\) 0.008} & 0.552 {\tiny \(\pm\) 0.008} & \textbf{0.996} {\tiny \(\pm\) 0.001} & \textbf{0.926} {\tiny \(\pm\) 0.009} & 0.994 {\tiny \(\pm\) 0.002} \\
    VAE+Agg & \textbf{0.871} {\tiny \(\pm\) 0.005} & 0.795 {\tiny \(\pm\) 0.009} & \textbf{0.593} {\tiny \(\pm\) 0.012} & 0.943 {\tiny \(\pm\) 0.010} & 0.918 {\tiny \(\pm\) 0.002} & 0.974 {\tiny \(\pm\) 0.003} \\
    \hline
    \\[-2ex]
    AE+KMeans & 0.854	 {\tiny \(\pm\) 0.003} & 0.803 {\tiny \(\pm\) 0.001} & 0.553 {\tiny \(\pm\) 0.004} & 0.995{\tiny \(\pm\) 0.000} & 0.922	 {\tiny \(\pm\) 0.006} & \textbf{0.998} {\tiny \(\pm\) 0.000} \\
    VAE+KMeans & 0.866	 {\tiny \(\pm\) 0.003} & \textbf{0.852}	{\tiny \(\pm\) 0.006} & 0.589	 {\tiny \(\pm\) 0.006} & 0.921 {\tiny \(\pm\) 0.005} & 0.924 {\tiny \(\pm\) 0.003} & 0.986 {\tiny \(\pm\) 0.001} \\
    AE+Agg' & 0.557		 {\tiny \(\pm\) 0.007} & 0.296 {\tiny \(\pm\) 0.006} & 0.322 {\tiny \(\pm\) 0.004} & 0.929{\tiny \(\pm\) 0.004} & 0.702 {\tiny \(\pm\) 0.004} & 0.520 {\tiny \(\pm\) 0.027} \\
    VAE+Agg' & 0.457		 {\tiny \(\pm\) 0.008} &0.324	{\tiny \(\pm\) 0.004} & 0.302	 {\tiny \(\pm\) 0.004} & 0.752{\tiny \(\pm\) 0.012} & 0.594{\tiny \(\pm\) 0.009} & 0.438{\tiny \(\pm\) 0.006} \\
    \hline
\end{tabular}
\caption{Model results for different clustering algorithms (ARI \(\pm\) standard error of the mean).
Agg = the agglomerative clustering algorithm from \texttt{sklearn} with \texttt{metric} being ``euclidean" and \texttt{linkage} being``ward".
Agg' = the agglomerative clustering algorithm from \texttt{sklearn} with \texttt{metric} being ``euclidean" and \texttt{linkage} being ``complete".
KMeans = the K-means algorithm from \texttt{sklearn}.
These scores are obtained for the optimal noise level and number of samples found for Agg.}
\label{table2}
\end{table}

%% file: implementation_details.tex
\section{LNS Implementation Details}
\subsection{GG 
 Datasets}
 \label{app: gg dataset details}
Here we introduce the design of the GG datasets.
Each image of the datasets contains several parts including one or several (groups of) objects and a background.
In the training dataset, the pixel values belonging to the same part co-vary with each other in the same channel while the pixel values belonging to different parts take independent values in the same channels.
Such a scheme is expected to abstract, at a high level, the statistical structure of realistic visual scenes, which is based on basic optics:
\begin{equation}
    L = I \times R,
\end{equation}
where $L$ is luminance, $I$ is illuminance, and $R$ is reflectance.
The rationale behind the above-mentioned relationship of different pixels is explained below.

The images in the datasets all have $3$ channels. 
For simplicity, let us only consider one of the channels.
The pixel values of one channel are the luminances of the reflected light that has the same wavelength. 
If two pixels belong to the same object, the ratio between their reflectances is fixed. 
This is because, in real life, the reflectance ratio between the different parts of the same object is fixed, as it is determined by the material.
On the other hand, the reflectances of two pixels belonging to two different objects that can appear in a visual scene are independent.
Furthermore, we assume that all the pixels of one object are exposed to the same lighting condition, namely the same illuminance, which is a reasonable simplifying assumption for simple visual scenes in real life.
With the above, one can draw the conclusion that the luminances (pixel values) of different pixels belonging to the same object should keep a fixed ratio between each other throughout the training dataset, the luminances (pixel values) of different pixels belonging to different objects should be independent.
Moreover, the pixel values of the same pixel in different channels are independent. 
Because the object might be in different lighting conditions in different visual scenes, and the illuminances of different channels under different lighting conditions should be independent.

Here, we describe how the different GG datasets are constructed.
We encourage the reader to revisit Figures ~\ref{fig:examples} and ~\ref{fig:segmasks} while reading the following for a better understanding.

\begin{description}[itemsep=0pt]

\item[Kanizsa Squares] contains three parts: a square, four circles, and a background.
The four circles always have the same pixel values. 
Their relative positions are fixed in the testing dataset but not in the training dataset.
In the testing dataset, although the square and the background have the same color, the model is expected to separate these two parts since it is primed by the training samples to do so.
Such expectation is reflected in the segmentation masks (Figure~\ref{fig:segmasks}).

\item[Closure] contains two parts, one outlined square and a background.
In the testing dataset, the interior of the outlined square and the background share the same pixel values, and the outlines of the squares are sometimes occluded partially.
The pixel value ratio of the outline and the square interior is kept the same as in the training dataset.
The model is expected to perceive the squares together with the outlines.

\item[Continuity] contains two parts in the training dataset, a circle and a background. 
In the testing dataset, the circle is fragmented, and several randomly generated fragments are placed at arbitrary positions. 
The model should perceive the existence of a circle in the testing dataset despite these disturbances.

\item[Proximity] contains either $2$ parts or $4$ parts.
In the former case, it contains a background one group of $6$ squares with an interspace of $1$ pixels. 
In the latter case, it contains a background and $3$ groups of $2$ squares. 
The interspaces between different squares within the same group are $1$ pixel wide, and the interspaces between different groups are at least $3$ pixel wide so that the two cases cannot be confounded. 
In the testing dataset, all the squares have the same color. 
However, the model is expected to group different squares together correctly given the interspaces.

\item[Gradient Occlusion] contains three parts, one tall rectangle with homogeneous color, one wide rectangle with gradient color, and a background. 
In the testing dataset, the homogeneously-colored rectangle always occludes the gradient-colored rectangle. 
However, the model is expected to group together the pixels belonging to the gradient-color rectangle, even if they do not have the same pixel values and are sometimes separated by the homogeneously-colored rectangle.

\item[Illusory Occlusion] contains $2$ parts, one striped background, and one striped square. 
In the testing dataset, part of the background and the foreground object have the same pixel value. 
However, the model is expected to identify the foreground-background structure.

\end{description}

For code to generate the GG datasets, please refer to \texttt{create\_datasets.py} in our git repository: \hyperlink{https://github.com/ZhengqingUUU/LatentNoiseSegmentation}{https://github.com/ZhengqingUUU/LatentNoiseSegmentation}.
The training set in our experiments contained $30,000$ images, the validation set contained $300$ images that were not in the training set, and the test set contained $100$ images.

\subsection{Deep Neural Networks}
\label{appendix:implementation}
\begin{table}[h]
\centering
\begin{tabular}{lc}
\hline
\\[-2ex]
\textbf{Hyperparameter} & \textbf{Value} \\ \hline
\\[-2ex]
Optimizer & Adam \\ 
Learning rate & \(5e^{-5}\) \\ 
Adam betas & (0.9, 0.999) \\
Batch size & 64 \\ 
Model latent dim & 15(GG)/50(CelebA) \\
Training iterations & \(1.1 e^{6}\) \\ 
AE loss function & MSE \\
VAE loss function & GECO\textbf{*} \citep{rezende2018}] \\
GECO g\_goal & 0.0006(GG)/0.0125 (CelebA)\\ \hline
\end{tabular}
\caption{Autoencoder and VAE hyperparameters used in the experiments. \textbf{*} Adapted from the implementation by \citet{engelcke2022}.}
\label{tab:my_label}
\end{table}

\begin{table}[H]
\centering
\begin{tabular}{llcccccc}
\hline
\\[-2ex]
& \textbf{Layer} & \textbf{Output Dimension} & \textbf{Kernel size} & \textbf{Stride} & \textbf{Padding} & \textbf{Activation function} \\ 
\hline
\\[-2ex]
\multirow{8}{*}{\rotatebox[origin=c]{0}{\textbf{$\operatorname{Enc}$}}}
& Conv2d & 32 & 3 & 1 & 1 & ReLU \\
& Conv2d & 64 & 4 & 2 & 1 & ReLU \\
& Conv2d & 64 & 4 & 2 & 1 & ReLU \\
& Conv2d & 128 & 4 & 2 & 1 & ReLU \\
& Conv2d & 128 & 4 & 2 & 1 & ReLU \\
& Conv2d & 256 & 4 & 1 & 0 & ReLU \\
& Linear & 128 & - & - & - & ReLU \\
& Linear & 15 & - & - & - & - \\ \hline
\\[-2ex]
\multirow{1}{*}{\rotatebox[origin=c]{0}{\textbf{$\operatorname{Dec}$}}}
& \multicolumn{6}{c}{Inverse of the encoder} \\ \hline
\end{tabular}
\caption{Autoencoder and VAE architecture.}
\label{tab:ae_model}
\end{table}

%% file: umap.tex
\section{Clusterability of Representations}
\subsection{AEs, but not VAEs, Perceive the Wrong Number of Clusters}
\label{appendix:clustering}
In our Segmentation algorithm, the expected number of clusters is provided on an image-by-image basis as \textit{a priori} information for the model. 
In most cases, the expected number of clusters coincides with the actual clusters spontaneously formed by $\Delta \Tilde{\mathbf{X}}_i$, which can be qualitatively verified with a UMAP visualization \citep{mcinnes2018}.
However, there are exceptions where $\Delta \Tilde{\mathbf{X}}_i$ is not readily grouped into the expected numbers of clusters. 
We observe such exceptions in segmentation using AE on the GG Proximity dataset (when there only exists $1$ group of $6$ squares) and the GG Illusory Occlusion dataset. 
Here, we present the UMAP visualization for them using one example per dataset. 
Note that while not shown here, we observe that the qualitative pattern of the AE being unable to capture the expected number of clusters generally holds for these two cases.

\begin{figure*}[htbp]
    \centering
    \begin{subfigure}[b]{0.153\textwidth}
        \includegraphics[width=\textwidth]{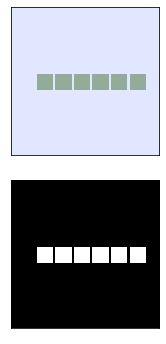}
        \caption{input (upper)\\target(lower)}
        \label{fig: ae proximity input target}
    \end{subfigure}
    \begin{subfigure}[b]{0.35\textwidth}
        \includegraphics[width=\textwidth]{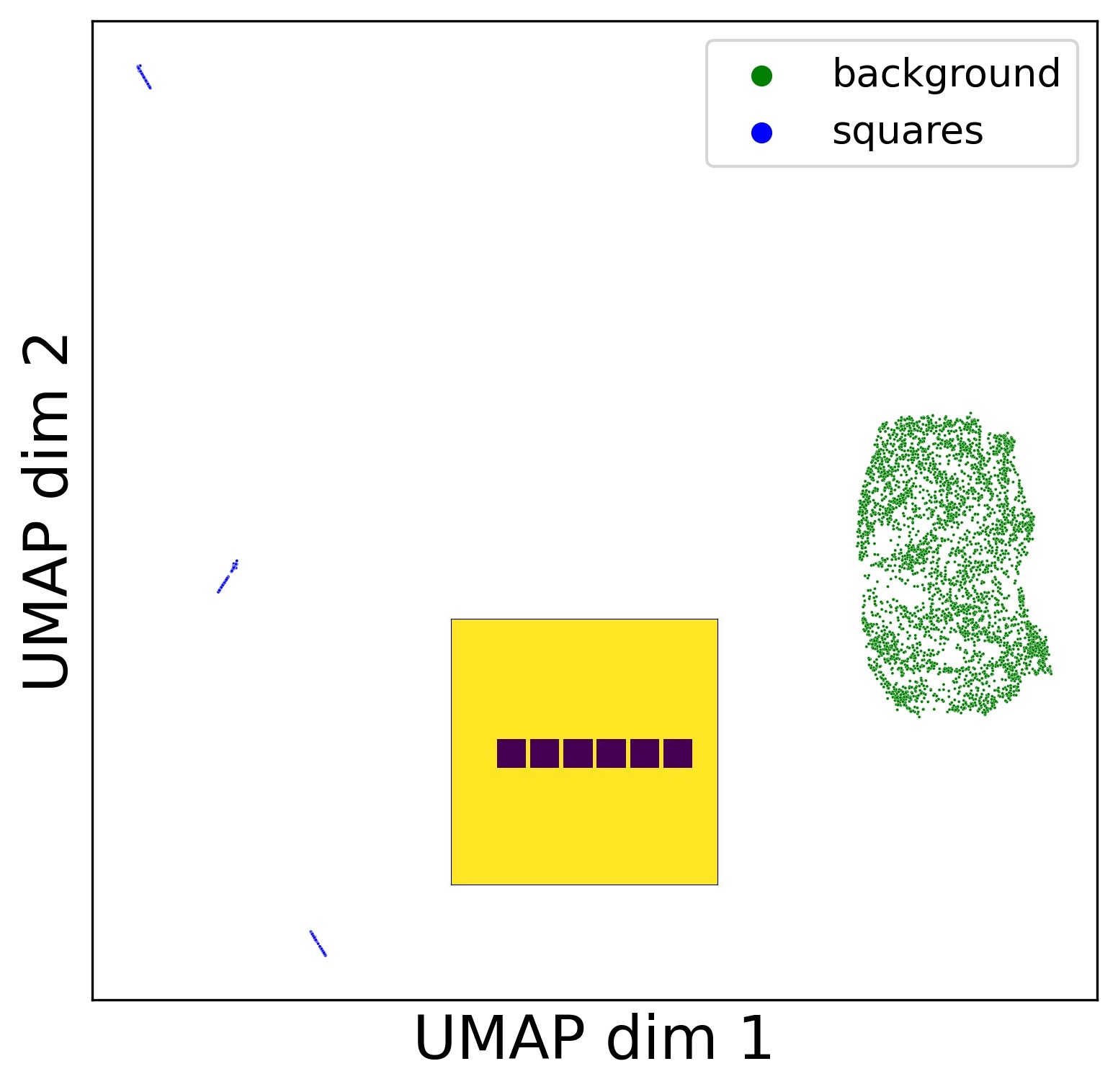}
        \caption{UMAP for AE}
        \label{fig: ae proximity umap}
    \end{subfigure}
    \begin{subfigure}[b]{0.35\textwidth}
        \includegraphics[width=\textwidth]{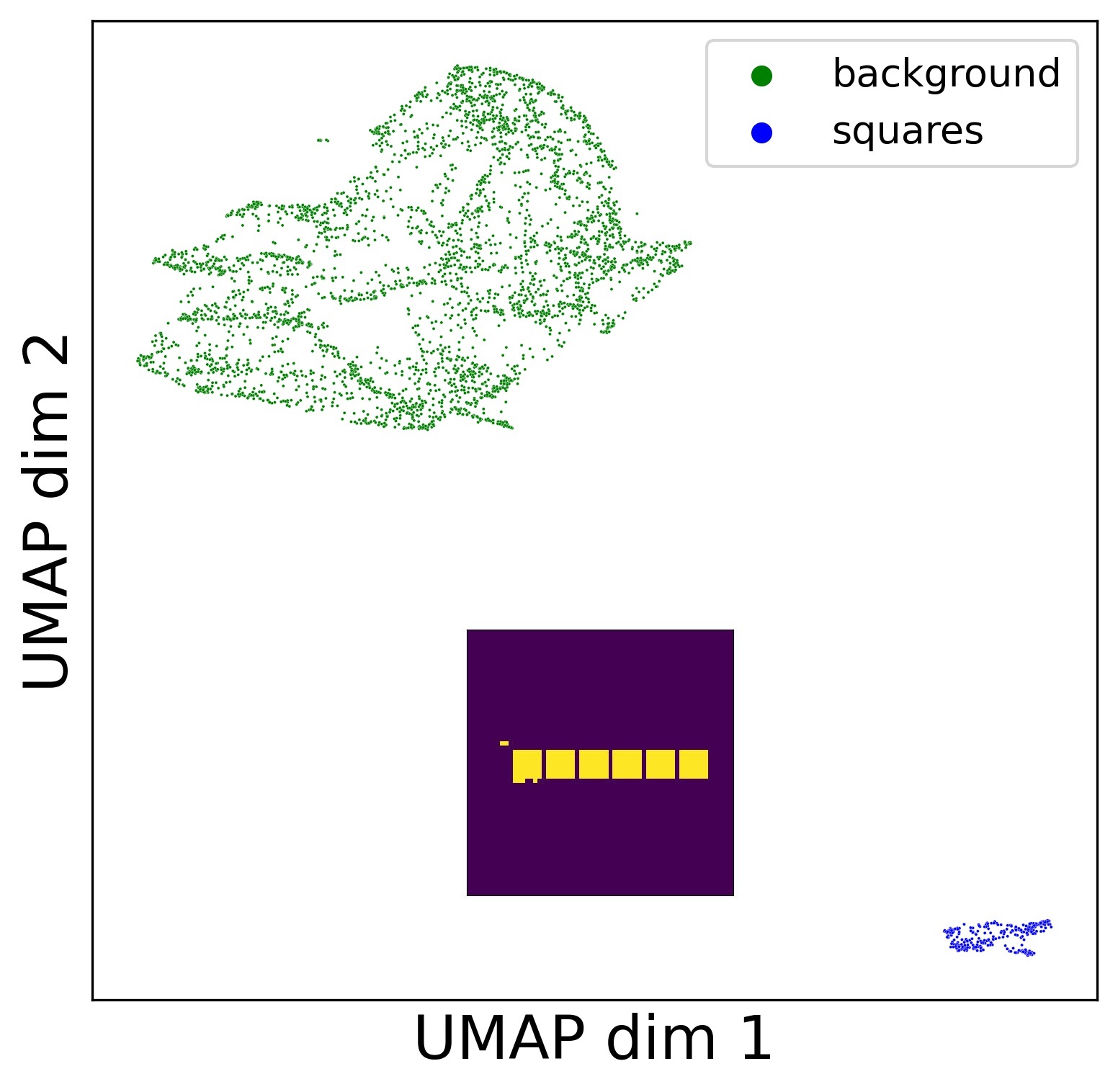}
        \caption{UMAP for VAE}
        \label{fig: vae proximity umap}
    \end{subfigure}
    \caption{The AE perceives the wrong numbers of clusters for the GG Proximity dataset with $1$ group of $6$ squares. (a) contains the input image and the expected mask. (b) and (c) show the visualization of $\Delta \Tilde{\mathbf{X}}^i$. Each dot corresponds to a pixel in the image. The visualization is made by using UMAP to reduce the $[C\times(N-1)]$-dimensional vector corresponding to each pixel to $2$ dimensions. 
    The color of the dots indicates which part the pixels correspond to. 
    The insets are the corresponding segmentation masks generated by the AE/VAE model.}
    \label{fig: ae proximity wrong cluster number}
\end{figure*}

In Figures ~\ref{fig: ae proximity umap} and ~\ref{fig: vae proximity umap}, we show the distribution of the vectors corresponding to each pixel in the $[C\times(N-1)]$-dimensional space where clustering is performed, when the models are confronted with an image from the GG Proximity dataset with $1$ group of $6$ squares.
The image and the expected target mask are shown in Figure~\ref{fig: ae proximity input target}.
In this case, the AE tends to separate the pixels belonging to the squares into $3$ clusters, perhaps as a result of being affected by training samples where there exist $3$ groups of $2$ squares; while the VAE tends to perceive the right number of clusters.
However, when given the expected number of clusters, which is $2$ in this case, the AE managed to nevertheless yield the correct segmentation mask (indicated by the inset in Figure~\ref{fig: ae proximity umap}).

\begin{figure*}[htbp]
    \centering
    \begin{subfigure}[b]{0.153\textwidth}
        \includegraphics[width=\textwidth]{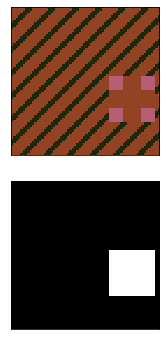}
        \caption{input (upper)\\target(lower)}
        \label{fig: ae illu_occlu input target}
    \end{subfigure}
    \begin{subfigure}[b]{0.35\textwidth}
        \includegraphics[width=\textwidth]{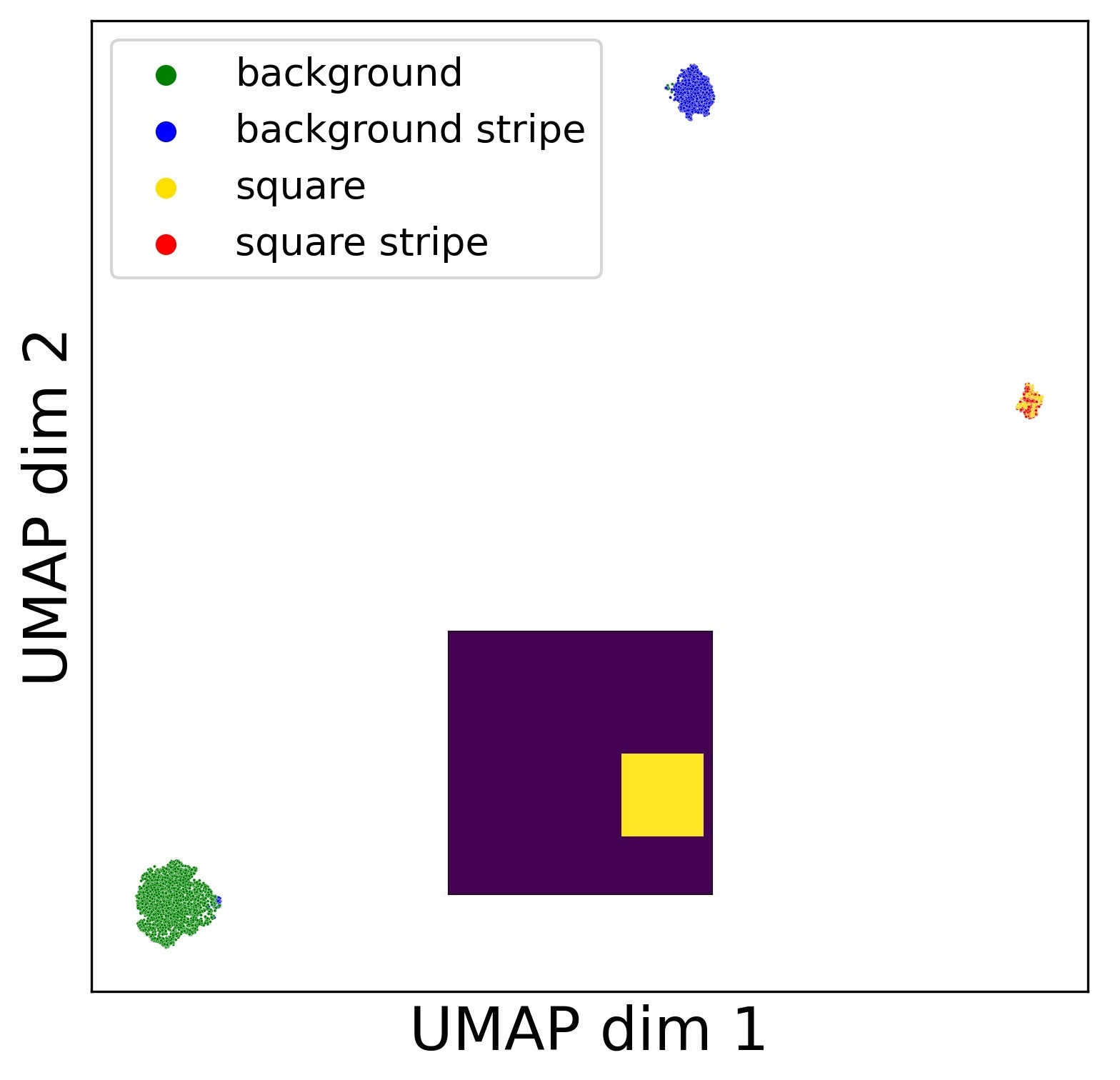}
        \caption{UMAP for AE}
        \label{fig: ae illu_occlu umap}
    \end{subfigure}
    \begin{subfigure}[b]{0.35\textwidth}
        \includegraphics[width=\textwidth]{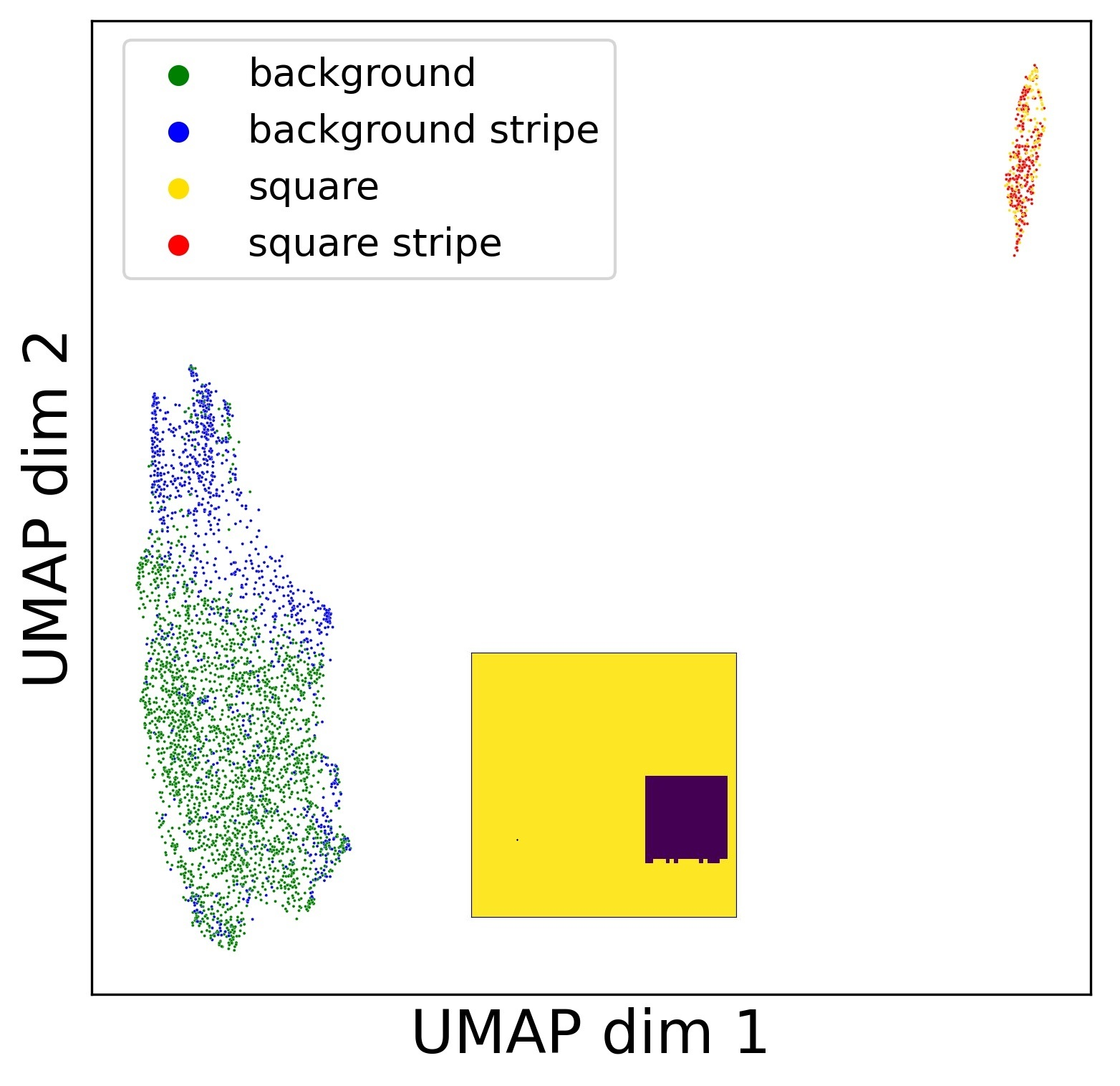}
        \caption{UMAP for VAE}
        \label{fig: vae illu_occlu umap}
    \end{subfigure}
    \caption{The AE perceives the wrong number of clusters for the GG Illusory Occlusion Dataset. Please refer to the caption of Figure~\ref{fig: ae proximity wrong cluster number}.}
    \label{fig: ae illu_occlu wrong cluster number}
\end{figure*}

A similar situation happens also for the GG Illusory Occlusion dataset. 
As shown in Figure~\ref{fig: ae illu_occlu wrong cluster number}, the AE tends to separate the background from the background stripes. 
We would expect these two parts to be grouped together, since they always co-vary with each other. 
Nonetheless, given the expected number of clusters, the AE manages to ultimately segment correctly.

\subsection{Latent Noise Segmentation Forms Clusterable Representations}
We further show the UMAP visualization of $\Delta \Tilde{\mathbf{X}}_i$ to demonstrate that LNS forms a meaningful clusterable representation of the pixels in general. 
$\Delta \Tilde{\mathbf{X}}_i$ is first reduced to $imag_x\times imag_y\times 2$ with UMAP performed on its last two dimensions, which contains $[C\times(N-1)]$-dimensional representations for each pixels. 
The resulting UMAP representation, accordingly, contains $2$-dimensional representations for all the pixels, which we visualize in Figure~\ref{fig: the big umap fig2}. 
The position of each dot shows the $2$-dimensional representation for each pixel, and we differentiate between different parts of the images with different colors.
The UMAP results in in Figure~\ref{fig: the big umap fig2} are generated while segmenting the samples in Figure~\ref{fig:results}.

\subsection{Model Test Reconstructions}
\begin{figure}[h]
\centering
\includegraphics[scale=0.35]{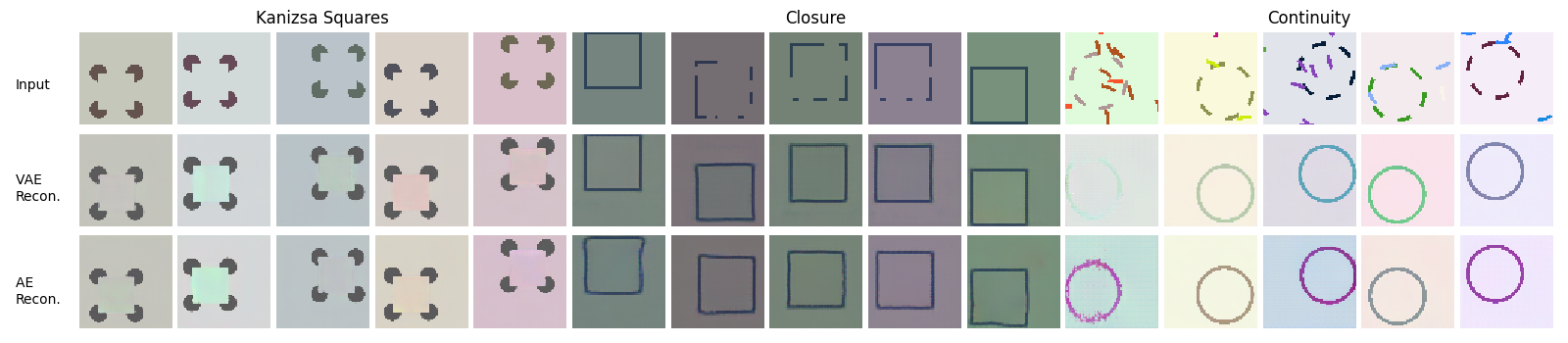}
\includegraphics[scale=0.35]{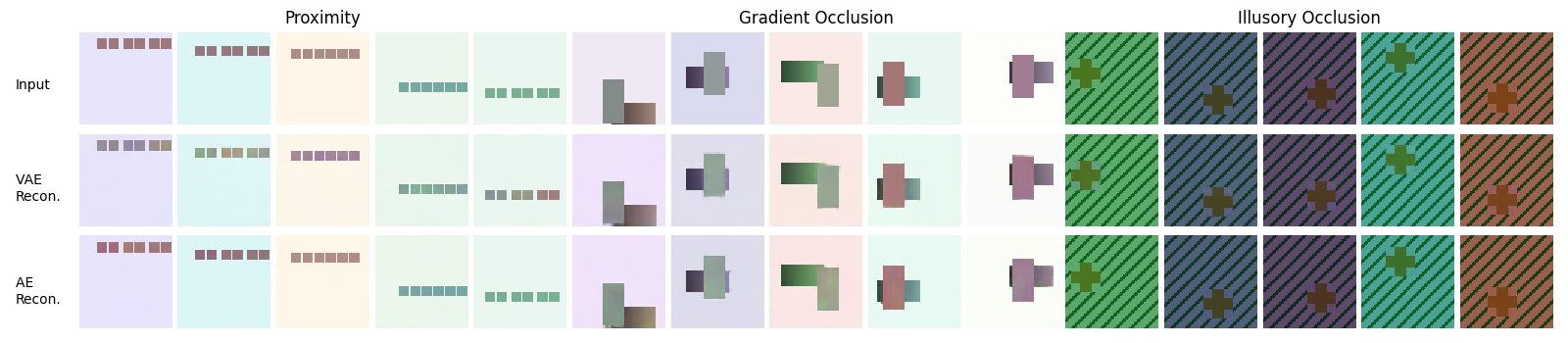}
\caption{\textbf{AE and VAE Model Reconstruction Examples.} The first row shows inputs, the second row shows VAE reconstructions, and the third row shows AE reconstructions. Randomly selected examples from the \textbf{Kanizsa Squares}, \textbf{Closure}, \textbf{Continuity}, \textbf{Proximity}, \textbf{Gradient Occlusion}, and \textbf{Illusory Occlusion} test datasets.}
\label{fig:model_recons}
\end{figure}

To support our control experiment, we also visualize the model output reconstructions (Figure \ref{fig:model_recons}).
In accordance with the control experiments, we find that the outputs in some cases contain some information about the identity of the objects, even when it does not veridically exist in the input image. 
However, as reported in Table \ref{table1}, the amount of information in the reconstructions is not as high as it is when using Latent Noise Segmentation.
This is because of the tension between reconstructing an image well (which would lead to poor segmentation without LNS in test samples), and segmenting the image components based on their color values. 
In other words, the fact that the model is still able to segment in some cases can be understood as an artefact caused by the small size of our model and imperfect reconstruction performance. 
The training process implicitly optimizes toward the success of LNS segmentation, but  against the success of direct reconstruction segmentation.

\newpage
\begin{figure}[H]
  \begin{subfigure}{\textwidth}
    \includegraphics[scale = 0.57, left]{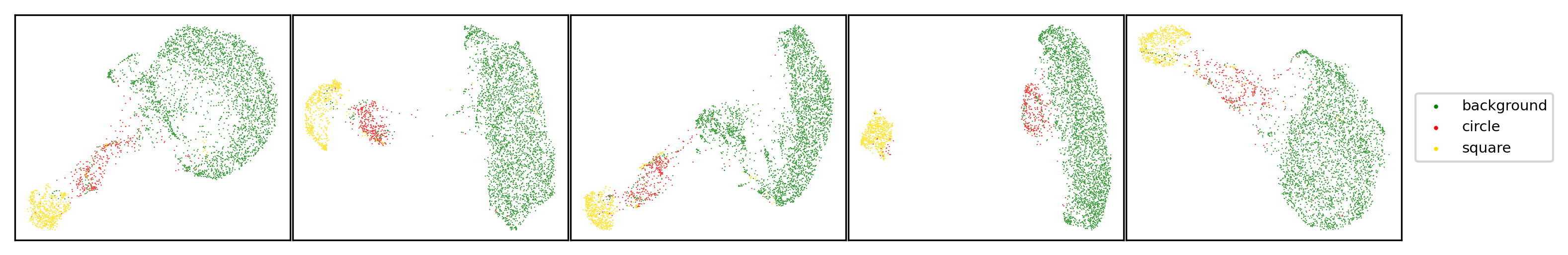}
    \caption{Model: VAE, dataset: Kanizsa Squares}
  \end{subfigure}

  \begin{subfigure}{\textwidth}
    \includegraphics[scale = 0.57, left]{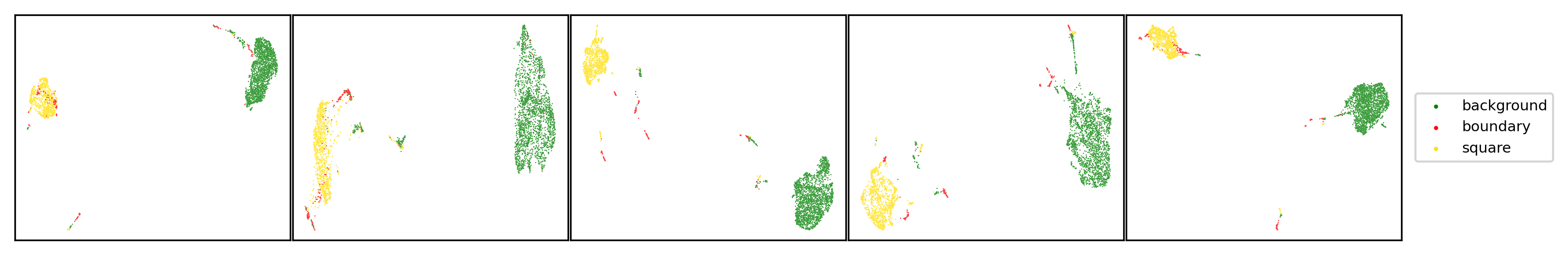}
    \caption{Model: VAE, dataset: Closure}
  \end{subfigure}

  \begin{subfigure}{\textwidth}
    \includegraphics[scale = 0.57, left]{figures/umap/UMAP_dShapeClosure_VAE.png}
    \caption{Model: VAE, dataset: Continuity}
  \end{subfigure}

    \begin{subfigure}{\textwidth}
    \includegraphics[scale = 0.57, left]{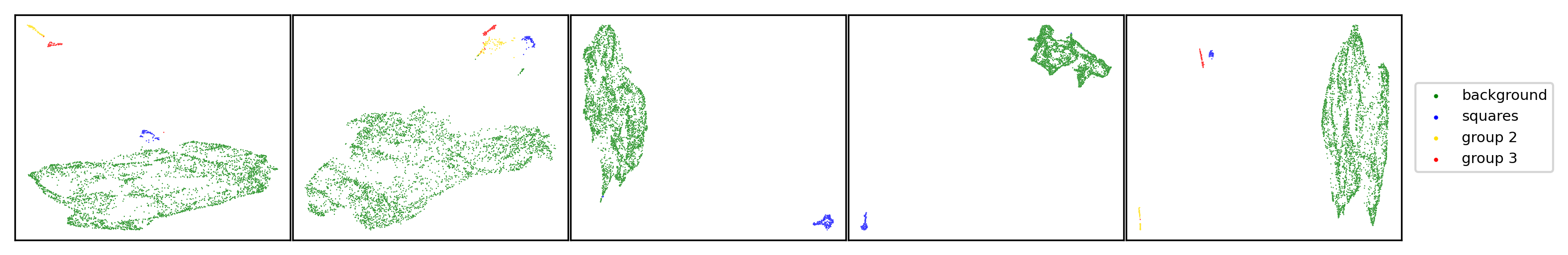}
    \caption{Model: VAE, dataset: Proximity}
  \end{subfigure}

    \begin{subfigure}{\textwidth}
    \includegraphics[scale = 0.57, left]{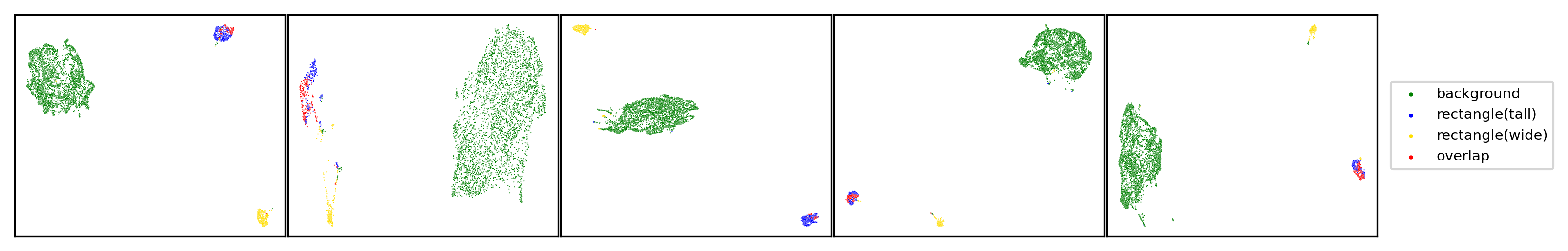}
    \caption{Model: VAE, dataset: Gradient Occlusion}
  \end{subfigure}

      \begin{subfigure}{\textwidth}
    \includegraphics[scale = 0.57, left]{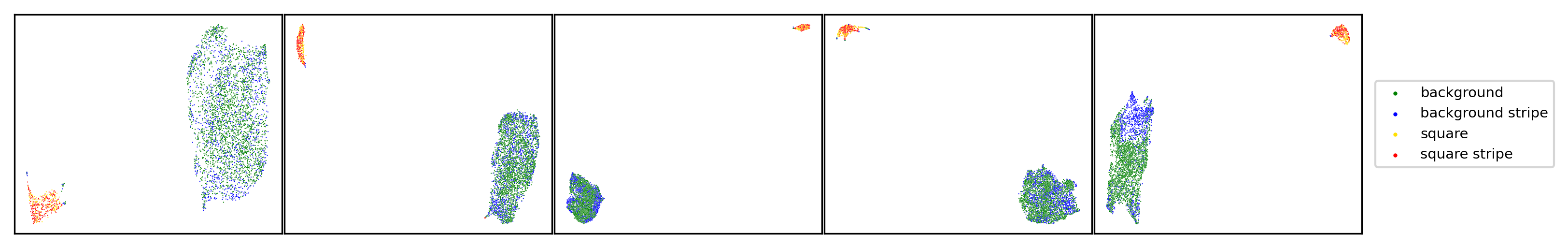}
    \caption{Model: VAE, dataset: Illusory Occlusion}
  \end{subfigure}

\label{fig: the big umap fig1}
  \end{figure}

\begin{figure}[H]
  \ContinuedFloat 
      \begin{subfigure}{\textwidth}
    \includegraphics[scale = 0.57, left]{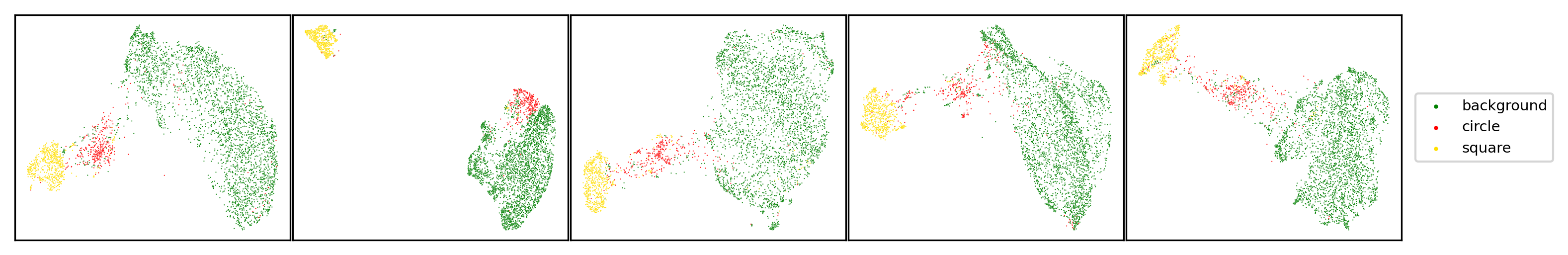}
    \caption{Model: AE, dataset: Kanizsa Squares}
  \end{subfigure}

  \begin{subfigure}{\textwidth}
    \includegraphics[scale = 0.57, left]{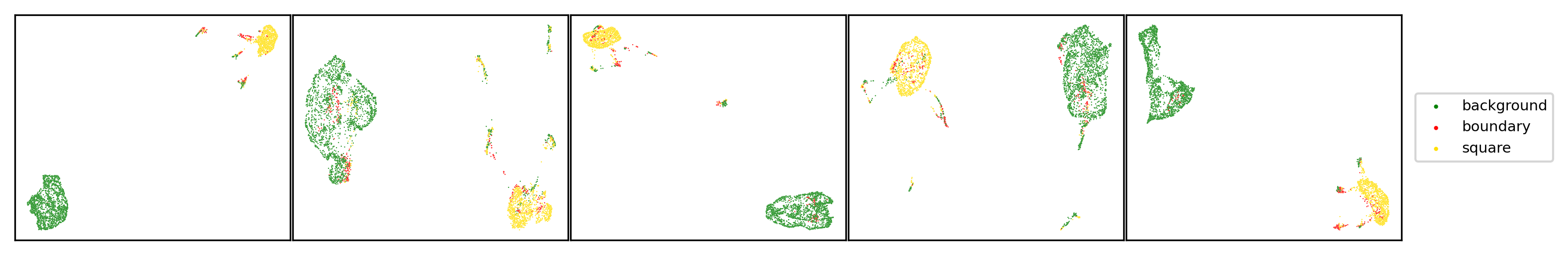}
    \caption{Model: AE, dataset: Closure}
  \end{subfigure}

  \begin{subfigure}{\textwidth}
    \includegraphics[scale = 0.57, left]{figures/umap/UMAP_dShapeClosure_AE.png}
    \caption{Model: AE, dataset: Continuity}
  \end{subfigure}

    \begin{subfigure}{\textwidth}
    \includegraphics[scale = 0.57, left]{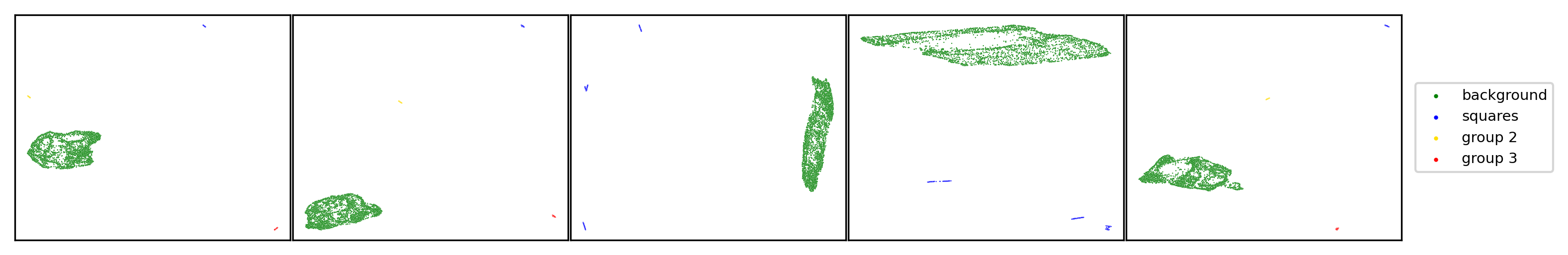}
    \caption{Model: AE, dataset: Proximity}
  \end{subfigure}

    \begin{subfigure}{\textwidth}
    \includegraphics[scale = 0.57, left]{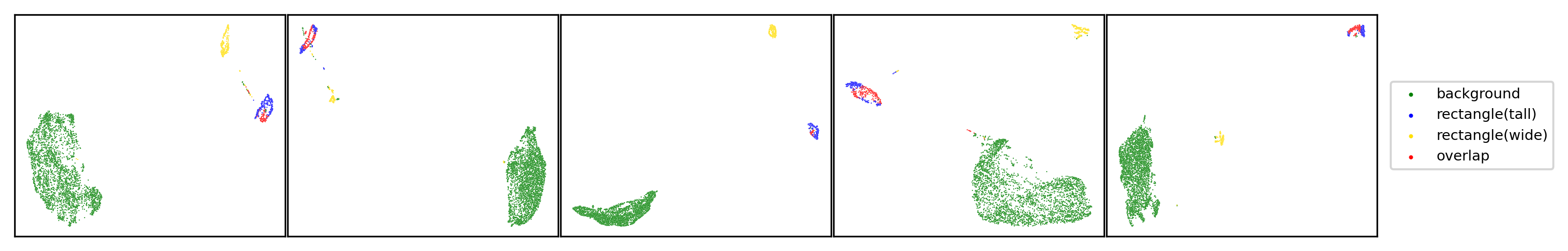}
    \caption{Model: AE, dataset: Gradient Occlusion}
  \end{subfigure}

      \begin{subfigure}{\textwidth}
    \includegraphics[scale = 0.57, left]{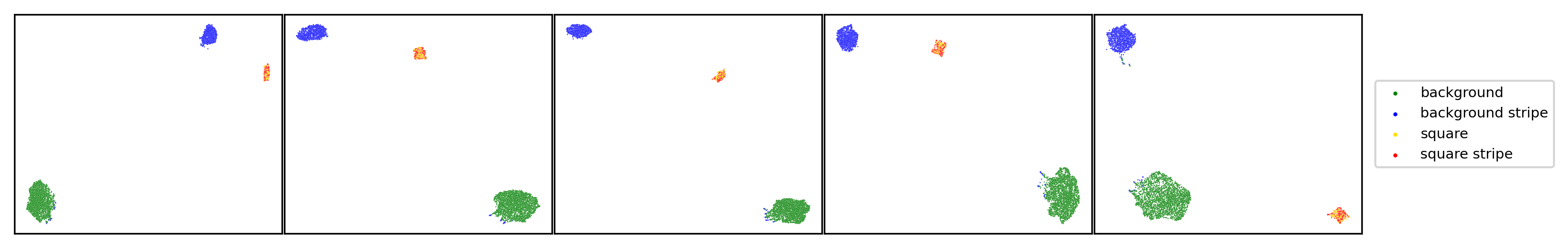}
    \caption{Model: AE, dataset: Illusory Occlusion}
  \end{subfigure}
  \caption{Visualization of $\Delta \Tilde{\mathbf{X}}^i$ with UMAP.}
   \label{fig: the big umap fig2}
\end{figure}

%% file: additional_outputs.tex
\subsection{Additional model outputs}
Here, we plot all model outputs for our testing set for a single seed for both model architectures (AE and VAE) to qualitatively confirm that our models capture the meaningful and relevant desired properties in our datasets, and that our qualitative model performance holds in general, and not only in a few hand-picked examples.
\label{appendix:additional_outputs}

\begin{figure}[H]
\begin{center}
%\framebox[4.0in]{$\;$}
\includegraphics[scale=0.25]{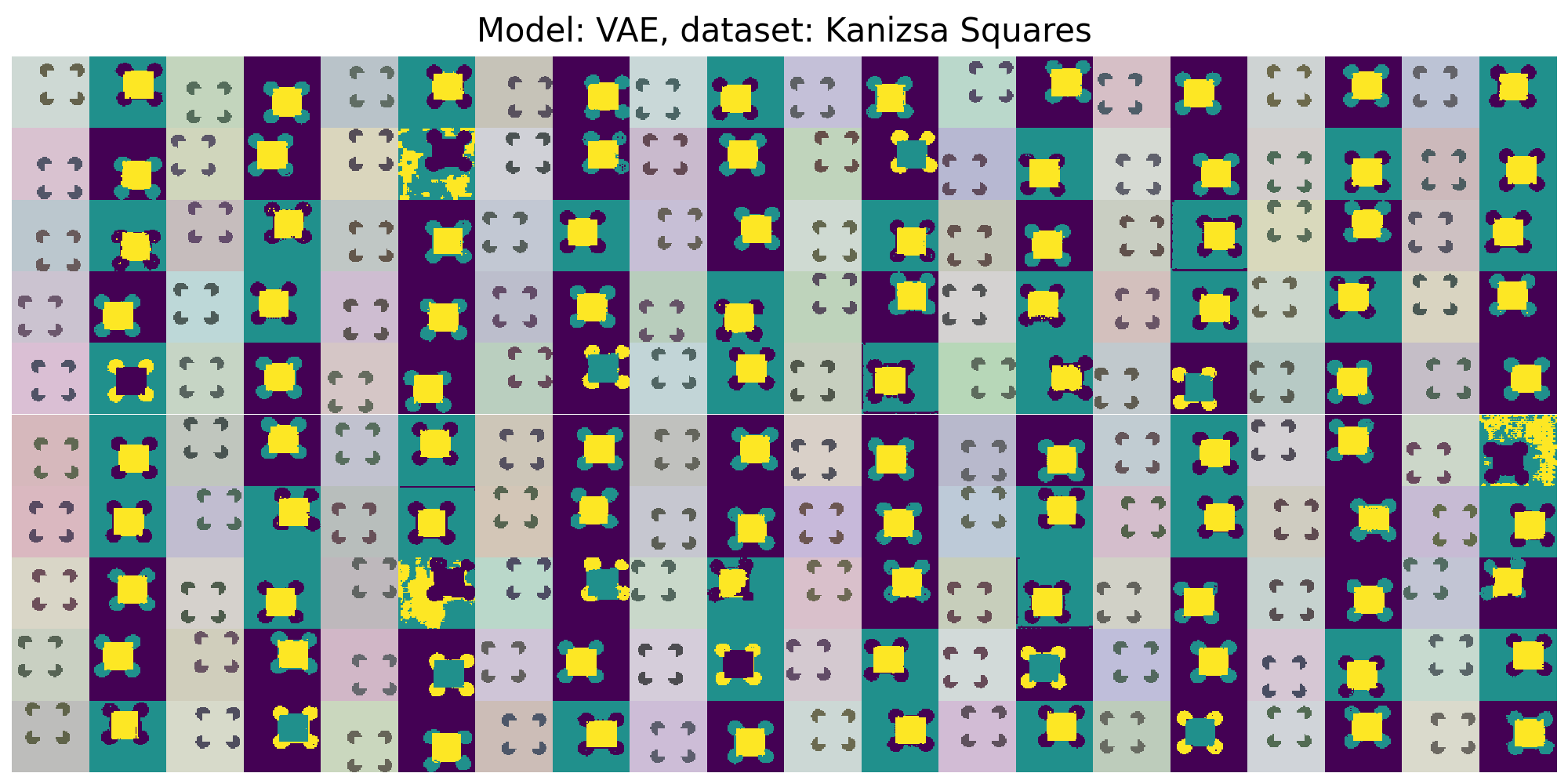}
\end{center}
\caption{\textbf{All Variational Autoencoder outputs for a single seed of the GG Kanizsa Squares dataset.} Inputs and outputs are displayed as horizontal pairs, with the input being the left image, and the model output being the right image.}
\label{fig:VAE_outputs_Kanizsa}
\end{figure}

\begin{figure}[H]
\begin{center}
%\framebox[4.0in]{$\;$}
\includegraphics[scale=0.25]{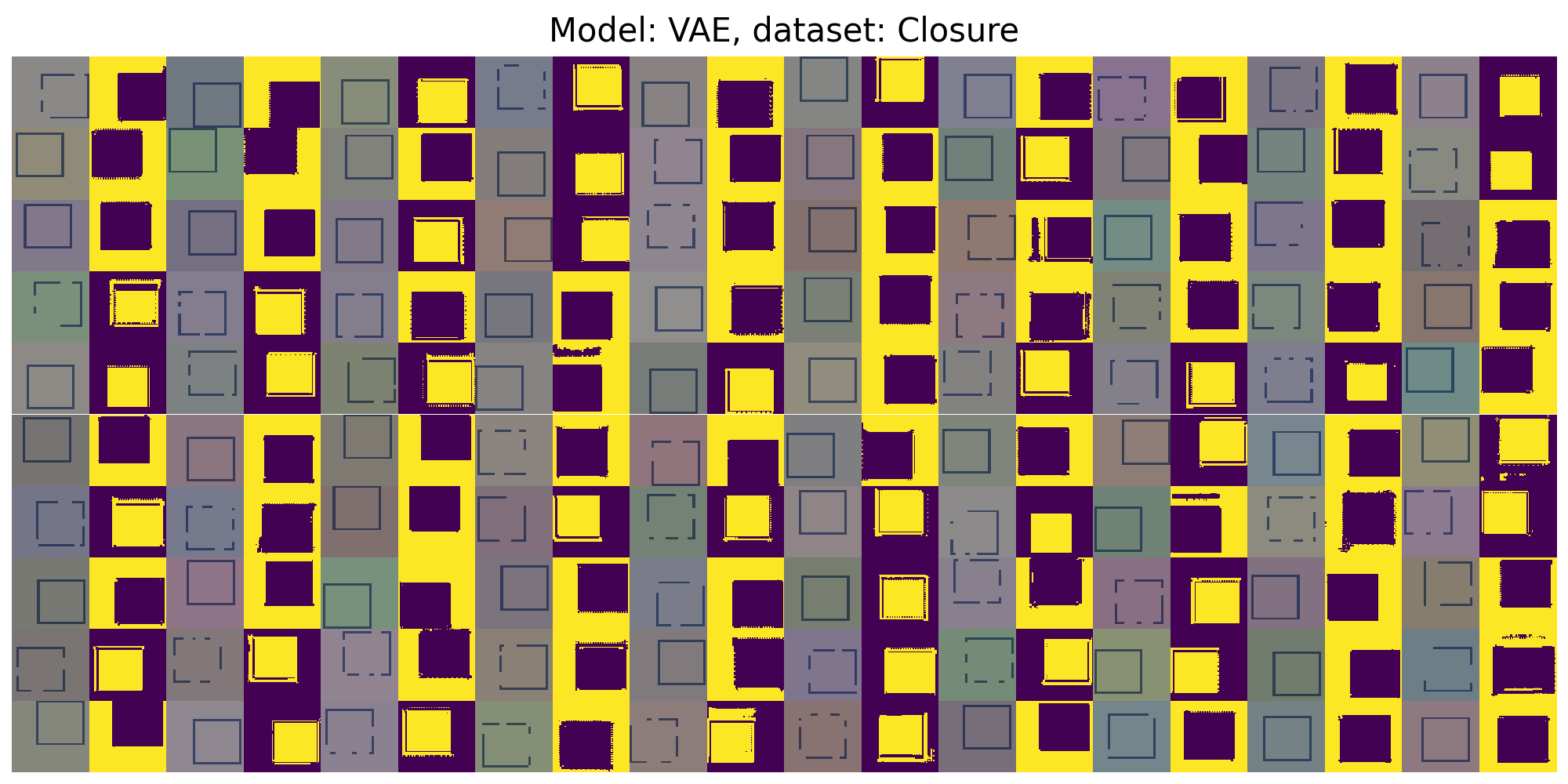}
\end{center}
\caption{\textbf{All Variational Autoencoder outputs for a single seed of the GG Closure dataset.} Inputs and outputs are displayed as horizontal pairs, with the input being the left image, and the model output being the right image.}
\label{fig:VAE_outputs_Closure}
\end{figure}

\begin{figure}[H]
\begin{center}
%\framebox[4.0in]{$\;$}
\includegraphics[scale=0.25]{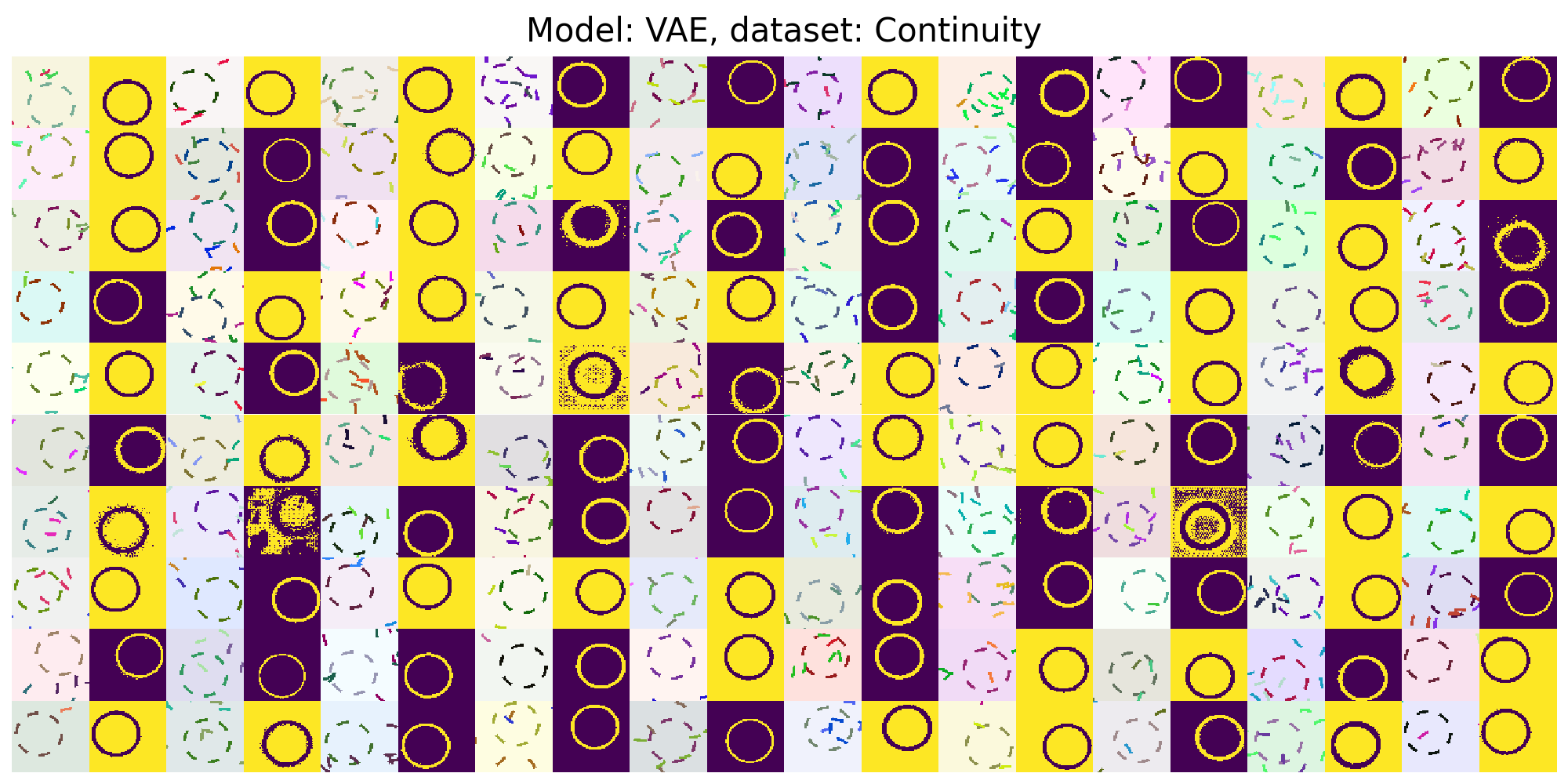}
\end{center}
\caption{\textbf{All Variational Autoencoder outputs for a single seed of the GG Continuity dataset.} Inputs and outputs are displayed as horizontal pairs, with the input being the left image, and the model output being the right image.}
\label{fig:VAE_outputs_Continuity}
\end{figure}

\begin{figure}[H]
\begin{center}
%\framebox[4.0in]{$\;$}
\includegraphics[scale=0.25]{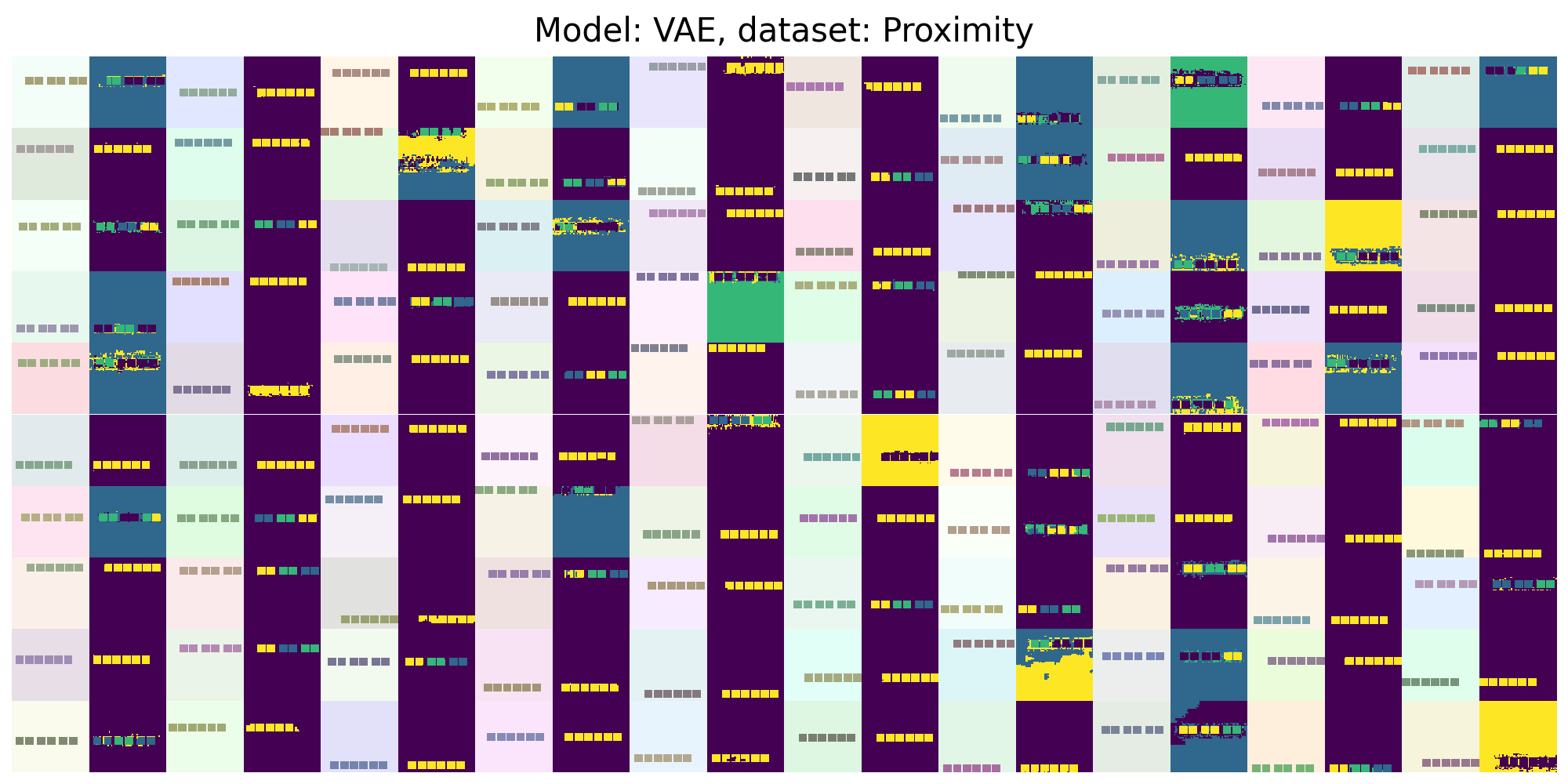}
\end{center}
\caption{\textbf{All Variational Autoencoder outputs for a single seed of the GG Proximity dataset.} Inputs and outputs are displayed as horizontal pairs, with the input being the left image, and the model output being the right image.}
\label{fig:VAE_outputs_Proximity}
\end{figure}

\begin{figure}[H]
\begin{center}
%\framebox[4.0in]{$\;$}
\includegraphics[scale=0.25]{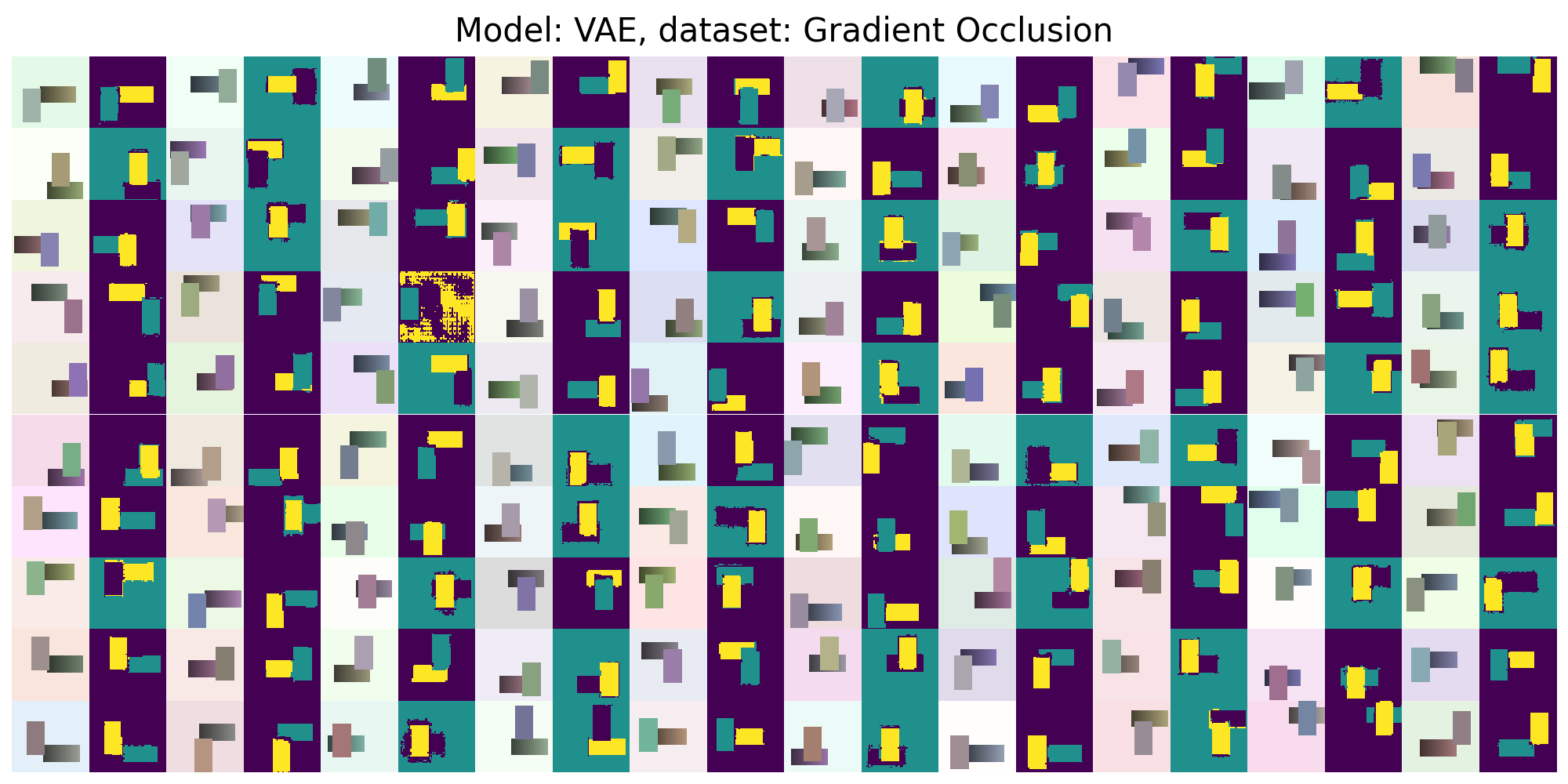}
\end{center}
\caption{\textbf{All Variational Autoencoder outputs for a single seed of the GG Gradient Occlusion dataset.} Inputs and outputs are displayed as horizontal pairs, with the input being the left image, and the model output being the right image.}
\label{fig:VAE_outputs_GradientOcclusion}
\end{figure}

\begin{figure}[H]
\begin{center}
%\framebox[4.0in]{$\;$}
\includegraphics[scale=0.25]{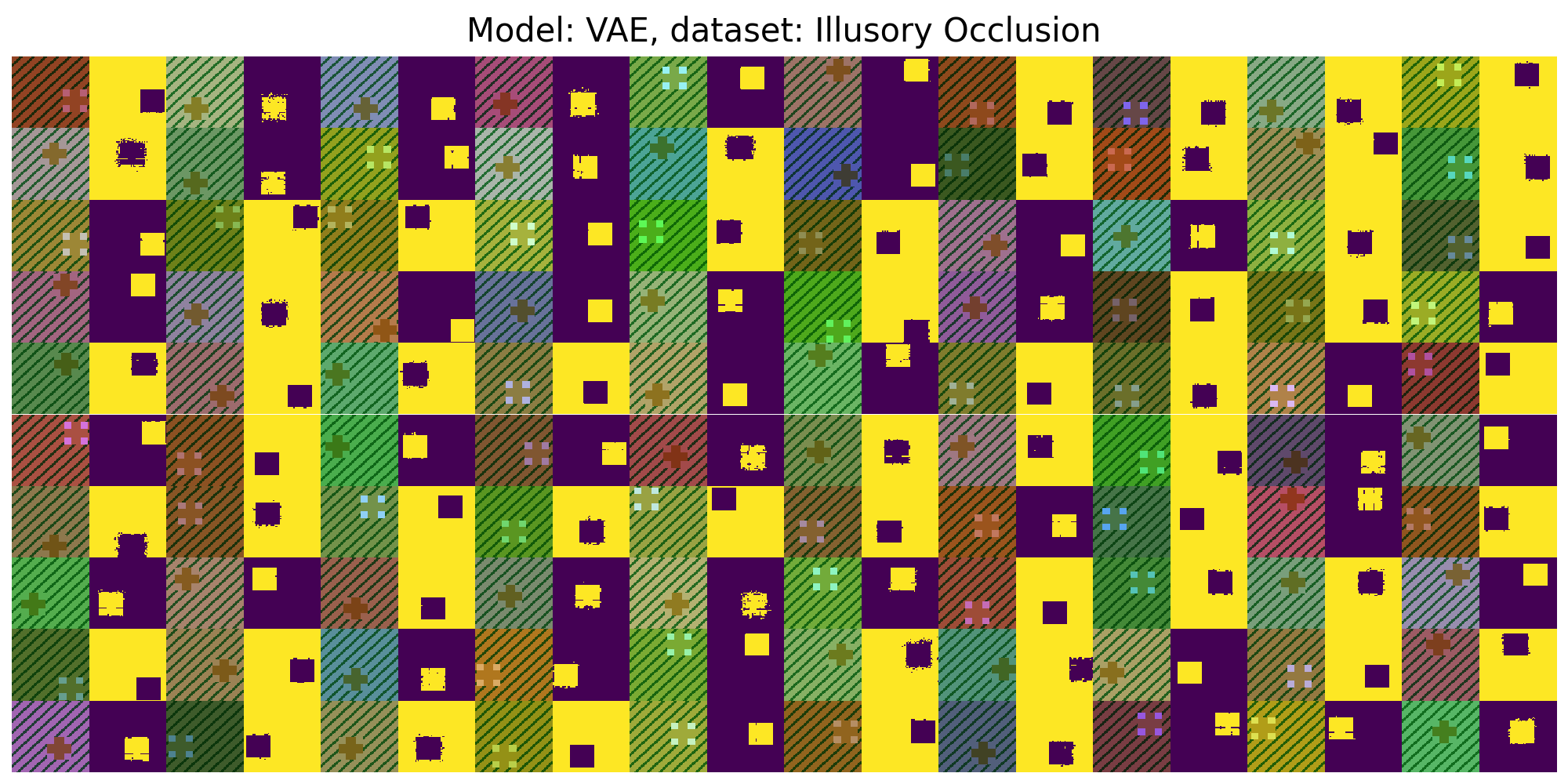}
\end{center}
\caption{\textbf{All Variational Autoencoder outputs for a single seed of the GG Illusory Occlusion dataset.} Inputs and outputs are displayed as horizontal pairs, with the input being the left image, and the model output being the right image.}
\label{fig:VAE_outputs_IllusoryOcclusion}
\end{figure}

\begin{figure}[h]
\begin{center}
%\framebox[4.0in]{$\;$}
\includegraphics[scale=0.25]{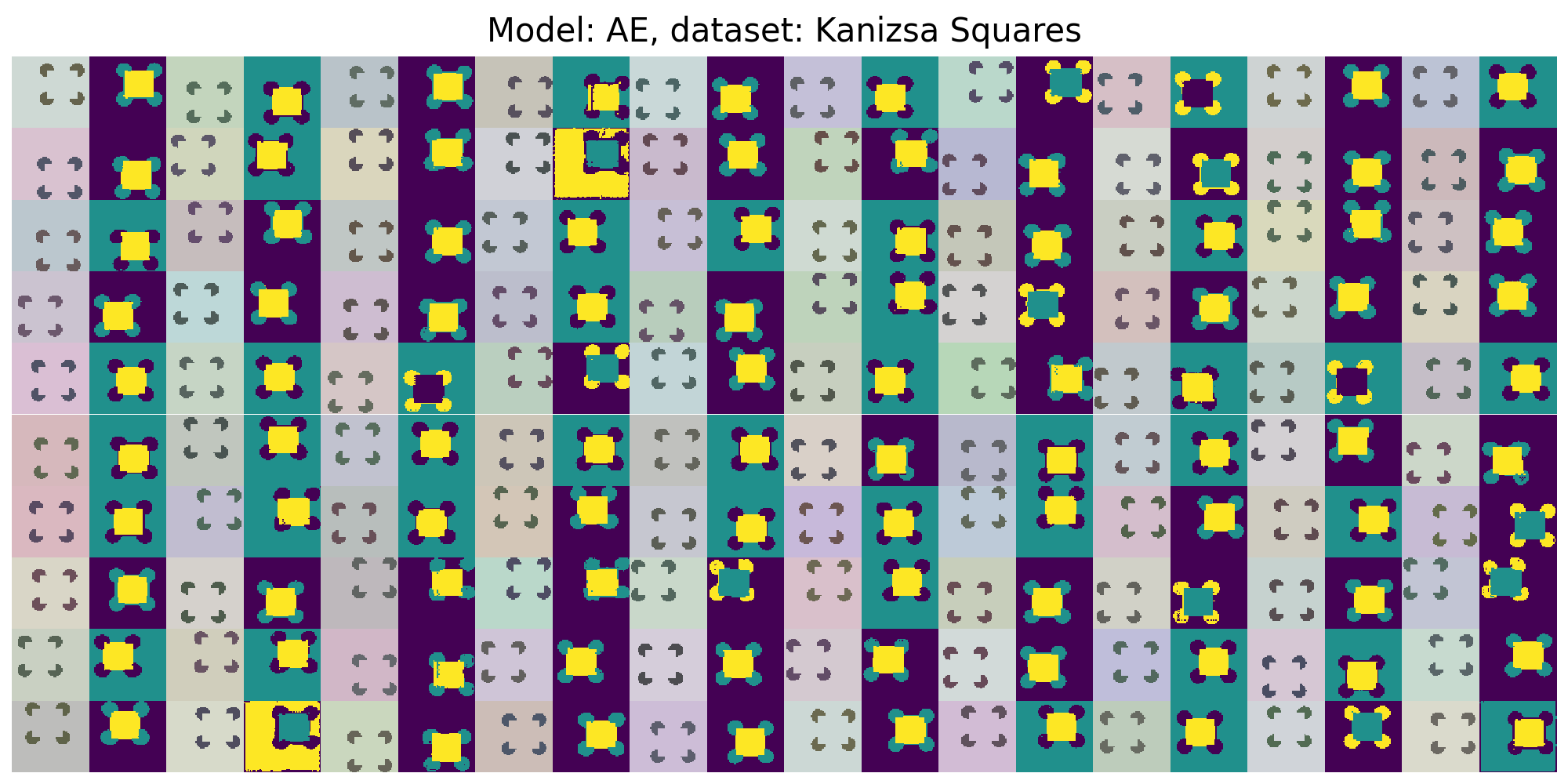}
\end{center}
\caption{\textbf{All Autoencoder outputs for a single seed of the GG Kanizsa Squares dataset.} Inputs and outputs are displayed as horizontal pairs, with the input being the left image, and the model output being the right image.}
\label{fig:AE_outputs_Kanizsa}
\end{figure}

\begin{figure}[H]
\begin{center}
%\framebox[4.0in]{$\;$}
\includegraphics[scale=0.25]{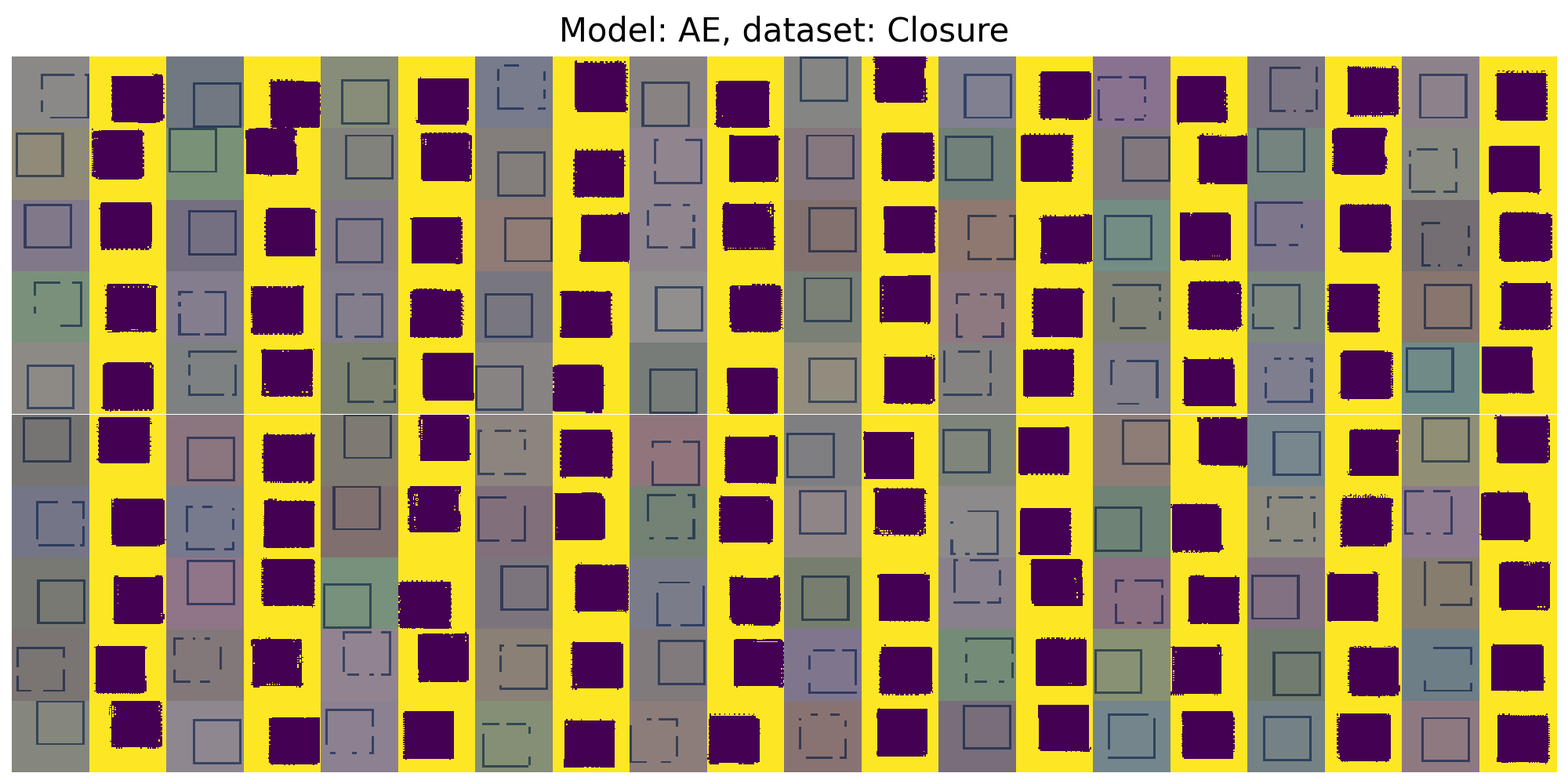}
\end{center}
\caption{\textbf{All Autoencoder outputs for a single seed of the GG Closure dataset.} Inputs and outputs are displayed as horizontal pairs, with the input being the left image, and the model output being the right image.}
\label{fig:AE_outputs_Closure}
\end{figure}

\begin{figure}[H]
\begin{center}
%\framebox[4.0in]{$\;$}
\includegraphics[scale=0.25]{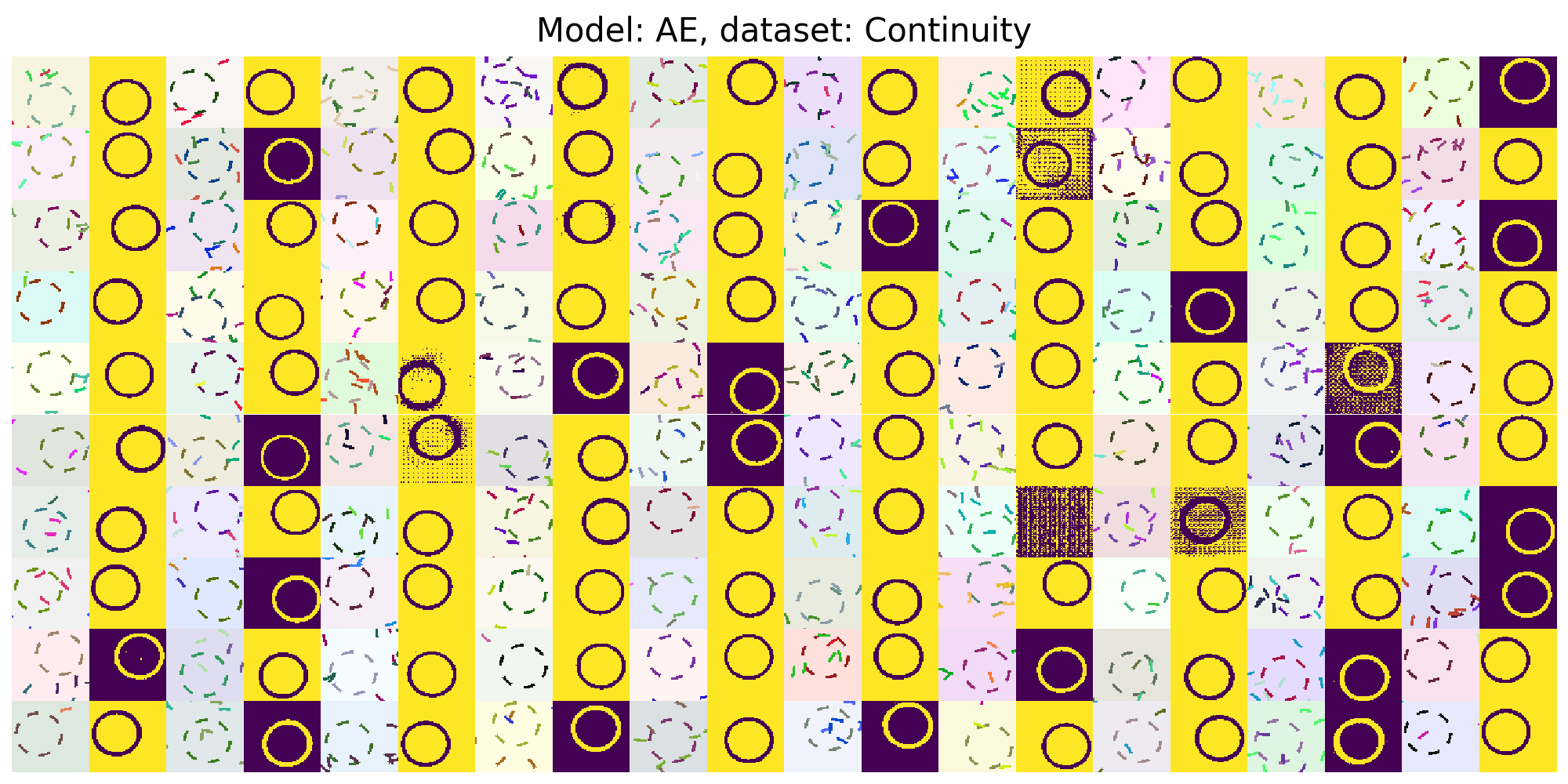}
\end{center}
\caption{\textbf{All Autoencoder outputs for a single seed of the GG Continuity dataset.} Inputs and outputs are displayed as horizontal pairs, with the input being the left image, and the model output being the right image.}
\label{fig:AE_outputs_Continuity}
\end{figure}

\begin{figure}[H]
\begin{center}
%\framebox[4.0in]{$\;$}
\includegraphics[scale=0.25]{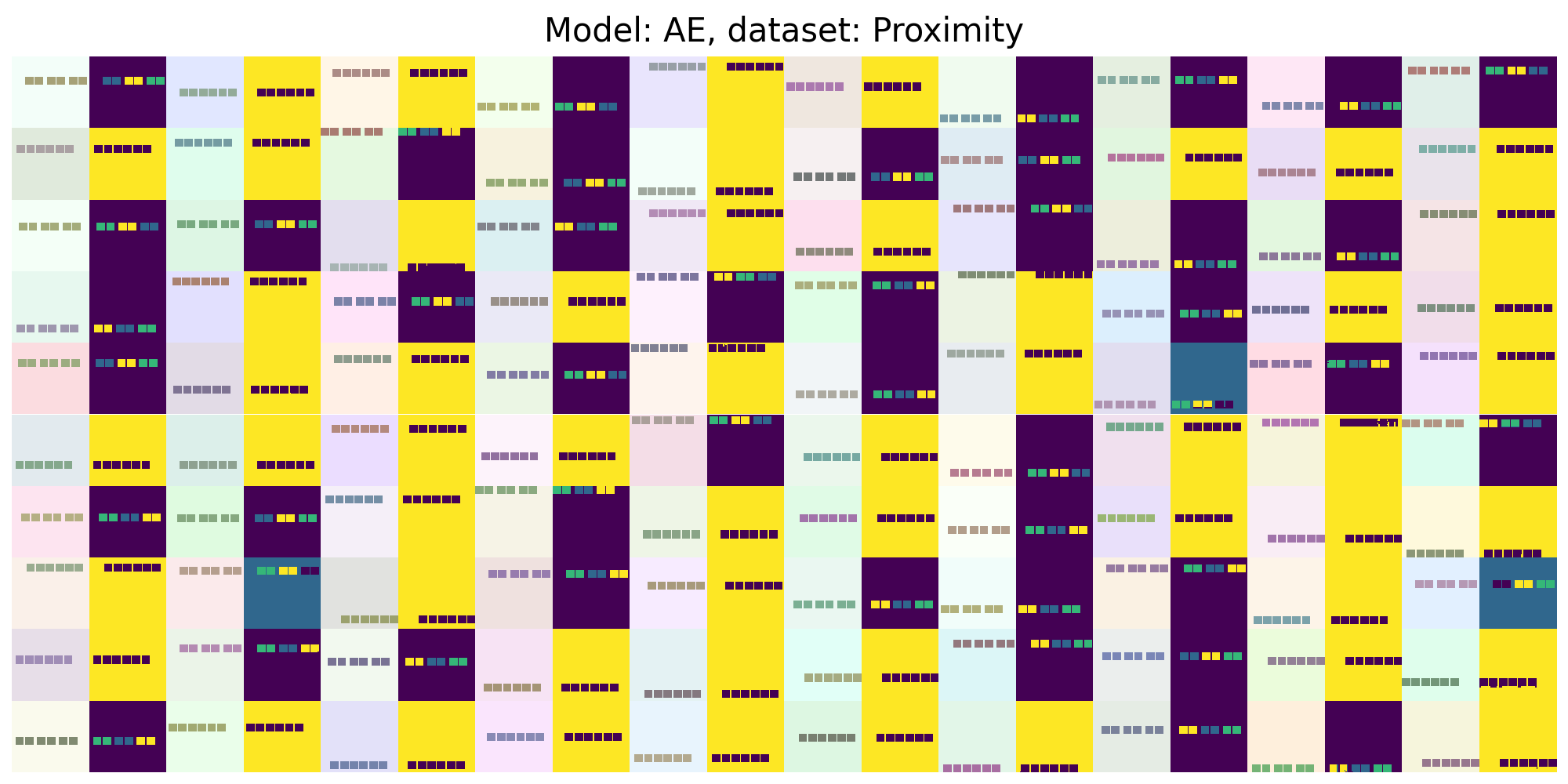}
\end{center}
\caption{\textbf{All Autoencoder outputs for a single seed of the GG Proximity dataset.} Inputs and outputs are displayed as horizontal pairs, with the input being the left image, and the model output being the right image.}
\label{fig:AE_outputs_Proximity}
\end{figure}

\begin{figure}[H]
\begin{center}
%\framebox[4.0in]{$\;$}
\includegraphics[scale=0.25]{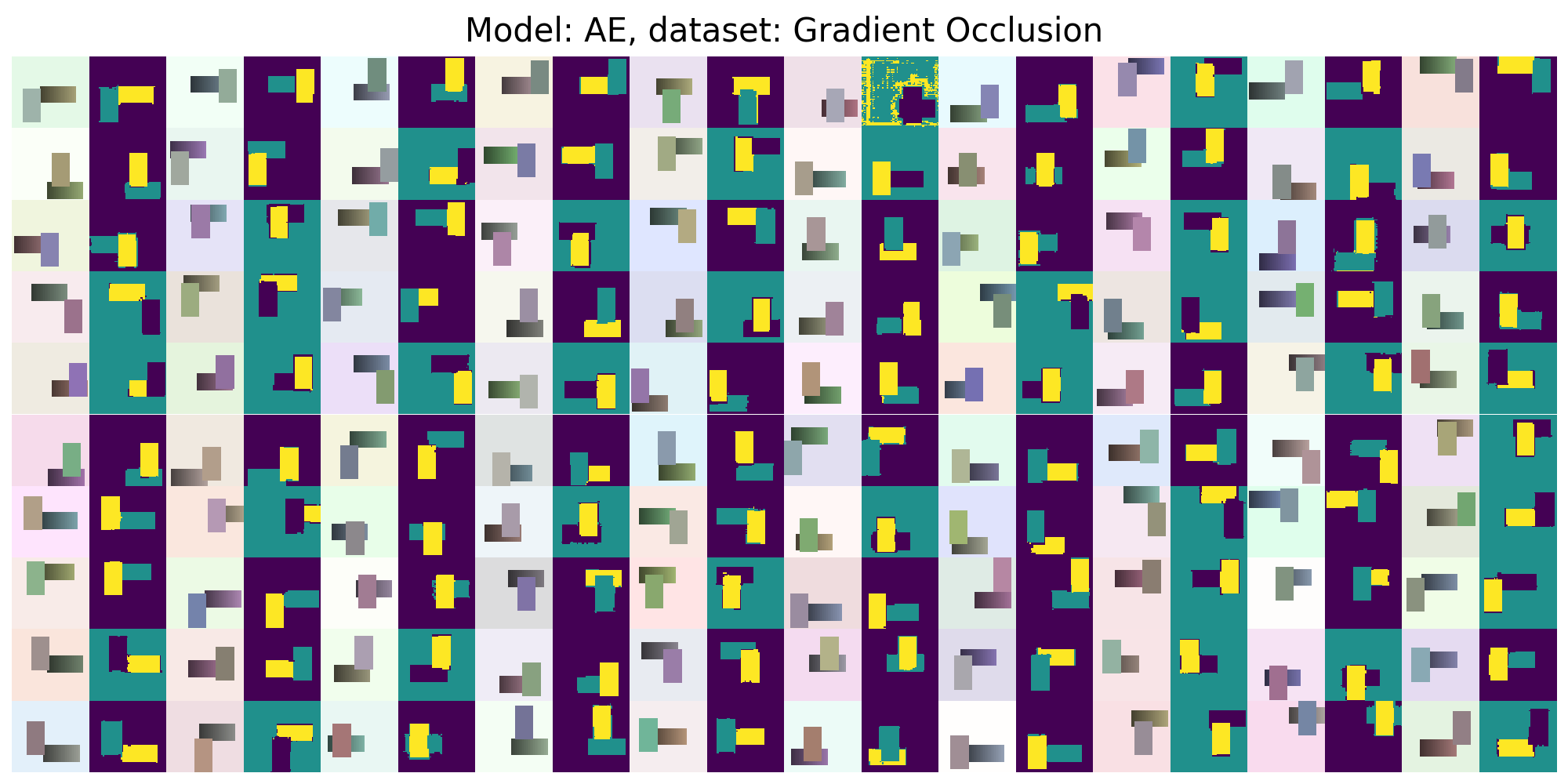}
\end{center}
\caption{\textbf{All Autoencoder outputs for a single seed of the GG Gradient Occlusion dataset.} Inputs and outputs are displayed as horizontal pairs, with the input being the left image, and the model output being the right image.}
\label{fig:AE_outputs_GradientOcclusion}
\end{figure}

\begin{figure}[H]
\begin{center}
%\framebox[4.0in]{$\;$}
\includegraphics[scale=0.25]{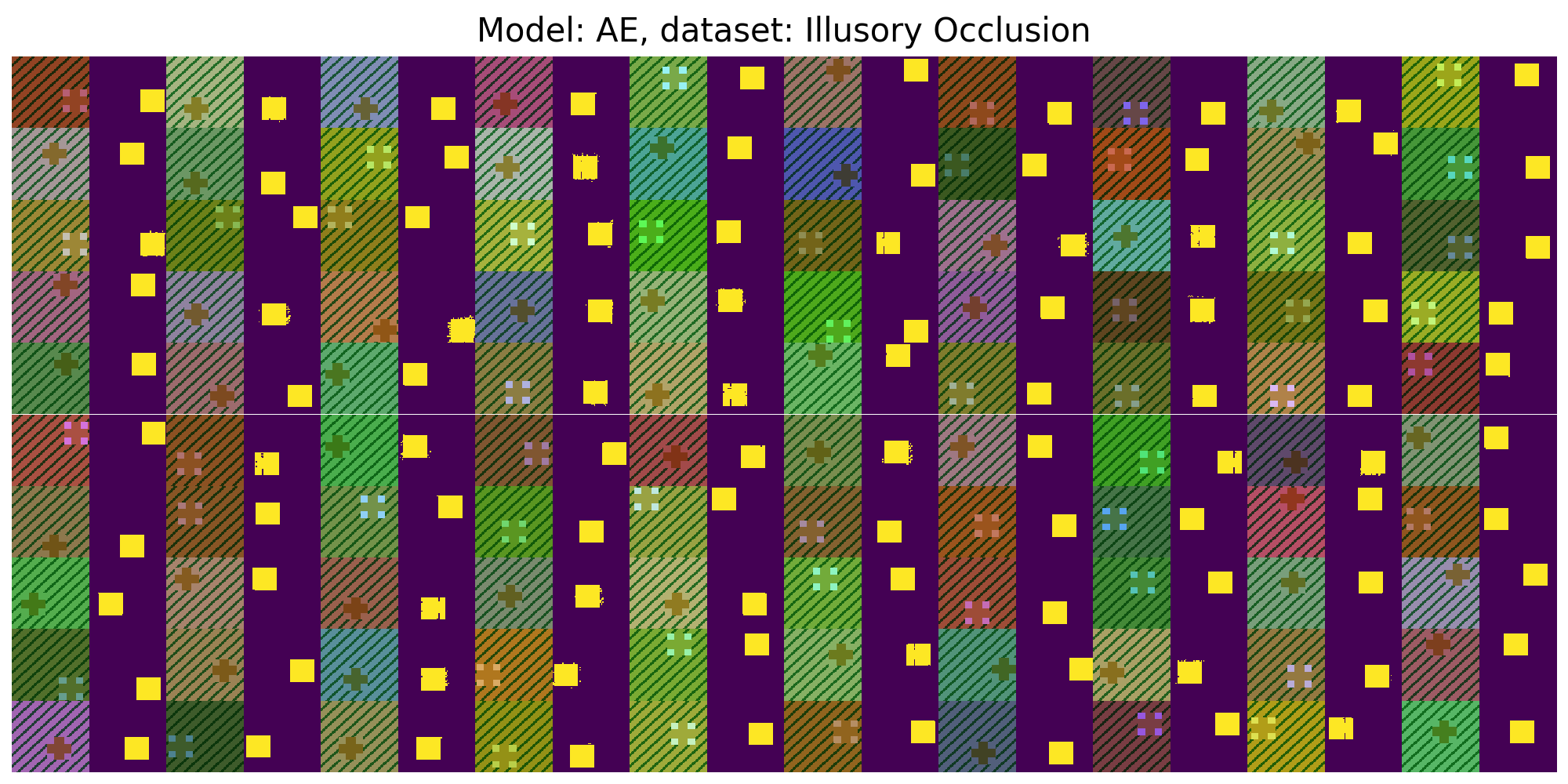}
\end{center}
\caption{\textbf{All Autoencoder outputs for a single seed of the GG Illusory Occlusion dataset.} Inputs and outputs are displayed as horizontal pairs, with the input being the left image, and the model output being the right image.}
\label{fig:AE_outputs_IllusoryOcclusion}
\end{figure}